\documentclass[10pt,twocolumn,letterpaper]{article}

\usepackage{iccv}
\usepackage{times}
\usepackage{epsfig}
\usepackage{graphicx}
\usepackage{amsmath}
\usepackage{amssymb}
\usepackage{booktabs}
\usepackage{multirow}
\usepackage{dblfloatfix}
\usepackage{subfig}
\usepackage[font=small]{caption}
\usepackage[accsupp]{axessibility}  
\usepackage[dvipsnames]{xcolor}

\usepackage[pagebackref=true,breaklinks=true,letterpaper=true,colorlinks,bookmarks=false]{hyperref}

\usepackage{enumitem}
\usepackage{authblk}
\usepackage{dblfloatfix}

\makeatletter
\renewcommand\AB@affilsepx{, \protect\Affilfont}
\makeatother

\usepackage[capitalize]{cleveref}
\crefname{section}{Sec.}{Secs.}
\Crefname{section}{Section}{Sections}
\Crefname{table}{Table}{Tables}
\crefname{table}{Tab.}{Tabs.}

\iccvfinalcopy 

\def\iccvPaperID{11098} 

\begin{document}

\title{FashionNTM: Multi-turn Fashion Image Retrieval via Cascaded Memory}
\author[1]{Anwesan Pal\thanks{\scriptsize Work primarily done during internship at Amazon. Additional details are available at \url{https://sites.google.com/eng.ucsd.edu/fashionntm}.}}
\author[2]{Sahil Wadhwa}
\author[2]{Ayush Jaiswal}
\author[2]{Xu Zhang}
\author[2]{Yue Wu}
\author[2]{Rakesh Chada}
\author[2]{Pradeep Natarajan}
\author[1]{Henrik I. Christensen}
\affil[1]{UC San Diego}
\affil[2]{Amazon Alexa Natural Understanding}

\affil[ ]{\protect\\ \footnotesize \texttt{\{a2pal, hichristensen\}@ucsd.edu}, \texttt{\{sahilwa, ayujaisw, xzhnamz, wuayue, rakchada, natarap\}@amazon.com}}


\maketitle
\ificcvfinal\thispagestyle{empty}\fi
\begin{abstract}
    Multi-turn textual feedback-based fashion image retrieval focuses on a real-world setting, where users can iteratively provide information to refine retrieval results until they find an item that fits all their requirements. In this work, we present a novel memory-based method, called FashionNTM, for such a multi-turn system. Our framework incorporates a new Cascaded Memory Neural Turing Machine (CM-NTM) approach for implicit state management, thereby learning to integrate information across all past turns to retrieve new images, for a given turn. Unlike vanilla Neural Turing Machine (NTM), our CM-NTM operates on multiple inputs, which interact with their respective memories via individual read and write heads, to learn complex relationships. Extensive evaluation results show that our proposed method outperforms the previous state-of-the-art algorithm by 50.5\%, on Multi-turn FashionIQ \cite{yuan2021conversational} -- the only existing multi-turn fashion dataset currently, in addition to having a relative improvement of 12.6\% on Multi-turn Shoes -- an extension of the single-turn Shoes dataset \cite{berg2010automatic} that we created in this work. Further analysis of the model in a real-world interactive setting demonstrates two important capabilities of our model -- memory retention across turns, and agnosticity to turn order for non-contradictory feedback. Finally, user study results show that images retrieved by FashionNTM were favored by 83.1\% over other multi-turn models.
    
\end{abstract}
\vspace{-20pt}
\section{Introduction}

Image retrieval has been extensively studied in the computer vision community, both using classical approaches \cite{chopra2005learning, turk1991eigenfaces, jegou2008hamming} and recently, using learning-based techniques \cite{arandjelovic2016netvlad, gordo2016deep, noh2017large, 10.1007/978-3-319-46448-0_1, hou2021disentanglement}. Existing works can be grouped based on input queries considered -- from image-only queries, commonly known as Content-Based Image Retrieval (CBIR) \cite{liu2016deepfashion, masi2018deep, schroff2015facenet}, to attributes \cite{han2017automatic}, sketches \cite{radenovic2018deep}, and natural language \cite{li2011text, zhang2005user}. However, most of these methods do not incorporate interactive user feedback, which is necessary for a personalized task such as fashion retrieval.

\begin{figure}[t]
    \centering
  \includegraphics[width=\linewidth]{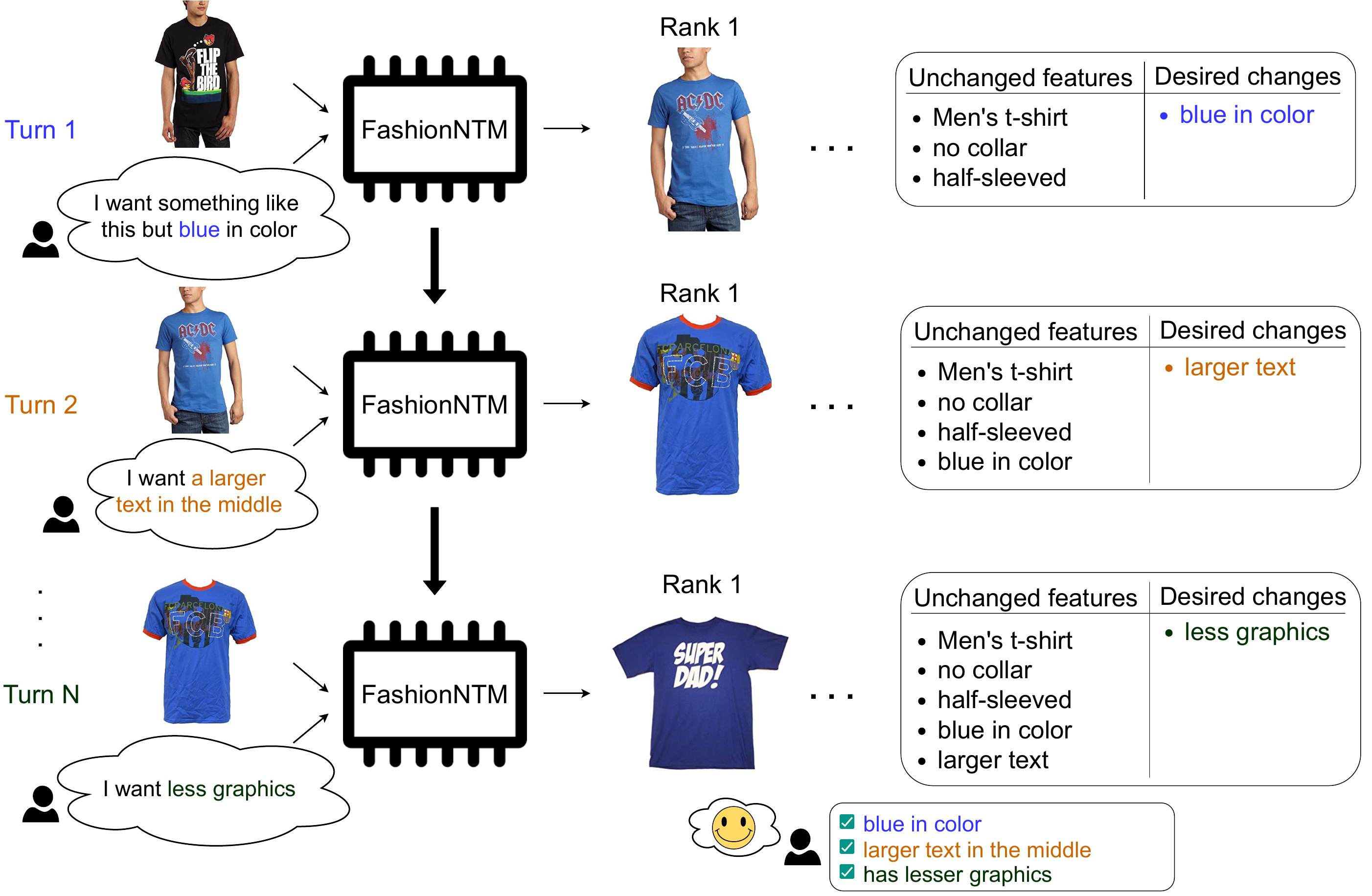}
  \caption{Illustration of multi-turn fashion image retrieval. Initially (\textcolor{blue}{Turn 1}), the system receives an initial query image, and a textual feedback mentioning the user's desired changes. The model then retrieves a ranked list of closest matching images. Subsequently, the user keeps refining their choice by providing more feedback, while the model retrieves newer images by considering both current and past feedback. This continues until the multi-turn system has successfully obtained the final retrieved image (\textcolor{OliveGreen}{Turn N}) with all the desired properties mentioned across every past turn.}
  \label{fig: intro}
  \vspace{-10pt}
\end{figure}

Textual feedback-based fashion image retrieval allows users to refine online shopping search results by providing information about how the results differ from their desired product (e.g., ``a dress like this but darker in color''). Several approaches for implementing such a system have been proposed recently \cite{Goenka_2022_CVPR, Chen_2020_CVPR, Lee_2021_CVPR, Wu_2021_CVPR, guo2018dialog, yuan2021conversational, zhang2019text}, which involve learning a joint representation via multi-modal information fusion across the query (reference) image and the associated feedback, and using it to retrieve the closest matching image in the database (product catalog) as the target.

Popular methods for fashion retrieval task \cite{Goenka_2022_CVPR, Lee_2021_CVPR, Chen_2020_CVPR, Baldrati_2022_CVPR, hou2021disentanglement} involve \textit{single-turn} exchange of information, where users provide feedback exactly once to update the search results. However, this is not characteristic of the real-world setting as online shopping customers typically start with a general idea of what they want and iteratively update the requirements until they find something that matches their desired features. This usually involves providing additional attributes, or modifying previously specified features in each turn to refine the search results. The ideal feedback-based fashion image retrieval system is, hence, inherently \textit{multi-turn}, as illustrated in \Cref{fig: intro}.

There are two major challenges associated with multi-turn image retrieval. First, there is a lack of sufficient training and evaluation datasets -- despite the abundance of single-turn fashion retrieval datasets, to the best of our knowledge, there is only one publicly available multi-turn fashion image retrieval dataset \cite{yuan2021conversational} currently. This is because labeling a sequence of images while ensuring continuity, consistency, and uni-directional information flow is a difficult problem. Thus, to facilitate research in this domain, we created a new fashion image retrieval dataset to allow for further benchmarking. Second, generalizing performance to real-world dynamic user interactive cases is non-trivial -- as this is still a relatively new research domain, most existing algorithms do not generalize beyond the training dataset to consider multiple turns of interactive feedback. In this work, we propose a novel memory-based framework to explicitly consider sequential feedback from users across multiple turns to retrieve desired items, both for the static image datasets, as well as real-world dynamic users.

Sequential modeling is a relatively mature field of research. However, a majority of the existing approaches \cite{rumelhart1985learning, jordan1997serial, 650093, hochreiter1997long, cho2014properties} do not maintain an explicit memory, and therefore cannot learn long and complex information. Vanilla memory network-based methods, which explicitly maintain an external memory, could be used for retaining past information, but they do not provide a robust mechanism to iteratively update their memory \cite{weston2014memory, sukhbaatar2015end}. In contrast, Neural Turing Machines (NTMs) \cite{graves2014neural} provide a fully differentiable model with sophisticated read and write operations to extract and update historical information in its explicit memory via an attention mechanism. 
Therefore, in this work, we build on NTMs to develop a novel framework for the multi-turn retrieval task. We further propose a novel Cascaded Memory Neural Turing Machine (CM-NTM) that allows us to encode multiple relationships from the features of a particular turn and store them over time across multiple memories in a multi-turn setting. This is similar to how multi-head attention (MHA) operates for transformers \cite{NIPS2017_3f5ee243}. To ensure that the individual memories effectively utilize each other's information, we link them together in a \textit{cascaded} fashion. Evaluation results demonstrate that our proposed approach improves the retrieval performance as compared to the previous state-of-the-art by $50.5\%$ on Multi-turn FashionIQ, and by $12.6\%$ on Multi-turn Shoes.

In summary, we make the following contributions. First, we propose a state-of-the-art memory-based framework, called FashionNTM, for multi-turn feedback-based fashion image retrieval, that uses an external memory to learn complex long-term relationships. Second, we develop a novel Cascaded Memory Neural Turing Machine (CM-NTM), that extends NTM to learn relationships across multiple inputs via additional controllers and read/write heads in a cascaded fashion. Third, we conduct experiments to show that the proposed approach outperforms existing state-of-the-art retrieval models by $50.5\%$ on Multi-turn FashionIQ \cite{yuan2021conversational}, and around $12.6\%$ on the multi-turn version of Shoes dataset \cite{berg2010automatic} respectively. Additionally, by performing an interactive analysis, we demonstrated two important capabilities of our multi-turn system -- memory retention across turns, and agnosticity to turn order for non-contradictory feedback. Finally, a user study result shows that on an average, the images retrieved by our model are preferred $83.1\%$ more than those from other multi-turn methods.
\section{Related Work}

\noindent\textbf{Single turn feedback-based fashion image retrieval} - Previous works in feedback-based fashion image retrieval have primarily focused on the single-turn scenario \cite{Goenka_2022_CVPR, mirchandani-etal-2022-fad, Lee_2021_CVPR, Chen_2020_CVPR, vo2019composing, Baldrati_2022_CVPR, 9857242, XuewenECCV20Fashion, chen2020learning, liu2021image, hosseinzadeh2020composed}, where a model is provided with a reference image along with an associated feedback text highlighting the desired attribute changes. The typical approach is to encode the multi-modal image and text input using pre-trained visual feature extractors \cite{he2016deep, howard2017mobilenets, lin2017feature} and sequential natural language processors \cite{hochreiter1997long, devlin2018bert}, respectively. More recently, pre-trained extractors such as Contrastive Language-Image Pre-training (CLIP) \cite{radford2021learning} have also been used \cite{Baldrati_2022_CVPR, 9857242}. This is then followed by a transformer-based decoder network \cite{NIPS2017_3f5ee243} for generating information-rich features for image retrieval from the database. Although these methods perform well in single-turn settings, they cannot be directly used for real-world applications that deal with multiple turns of information exchange.
\begin{figure*}[t]
    \centering
  \includegraphics[width=\linewidth]{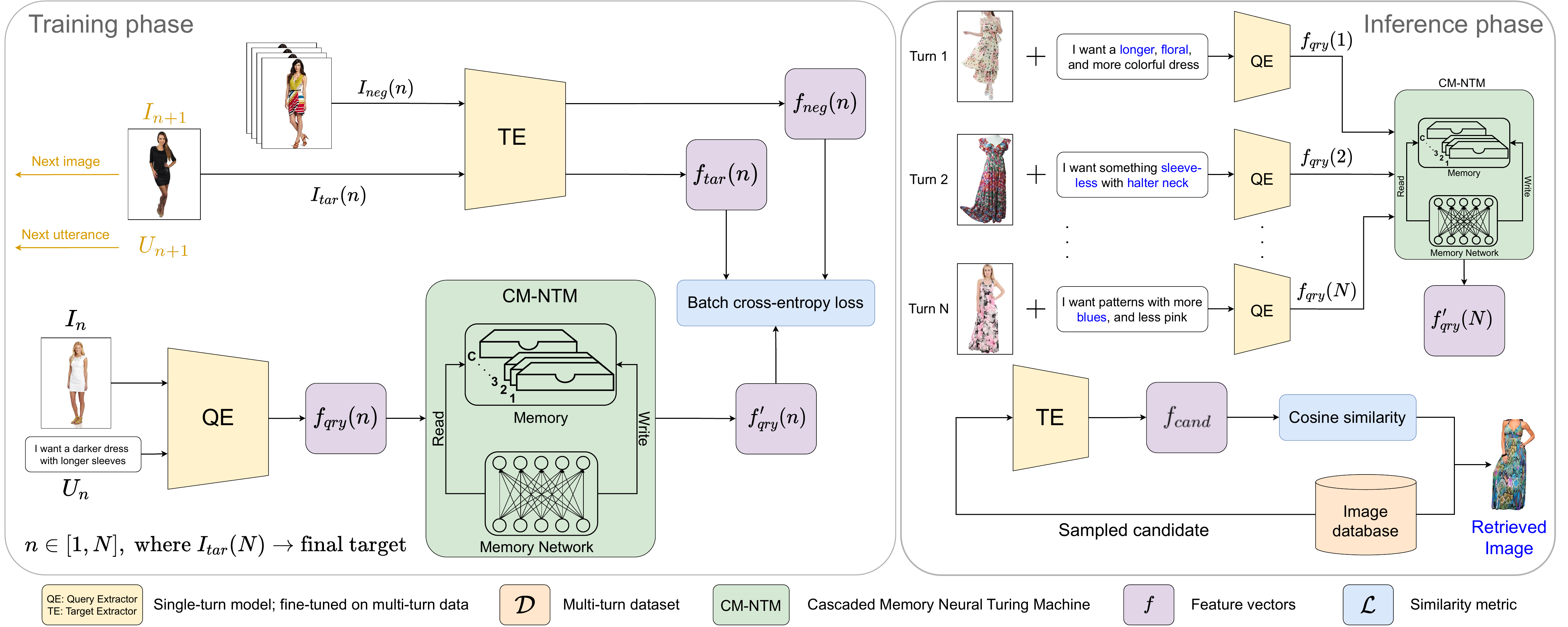}
  \caption{The complete FashionNTM framework. During training, the model receives an input query image, $I_n$, and associated feedback, $U_n$. From these, query features $f_{qry}(n)$ are computed using an off-the-shelf single-turn fashion image retrieval model. These features are then fed into the memory network which learns to retain useful information from the current turn, and combine it with information from previous turns by interacting with an external memory bank, via read/write operations. The memory-modified features $f'_{qry}(n)$ are compared with the target index features $f_{tar}(n)$, and other negative samples $f_{neg}(n)$, to compute the training loss. At inference, the model receives a series of multi-modal inputs turn-wise and computes the final modified feature $f'_{qry}(N)$ using the trained memory network. This is compared with features $f_{cand}$ derived from different candidate images in the database to retrieve the closest matching images.}
  \label{fig: overall_arch}
  \vspace{-10pt}
\end{figure*}

\noindent\textbf{Multi-turn visuo-linguistic methods} - A few methods have been proposed recently for fusing visual and textual input across multiple turns of information exchange. A popular application has been the video dialog task \cite{pham2022video, hori2019end, lin2019entropy, le2020bist, guo2018dialog}, where a trained system is asked to answer questions based on an ongoing video dialog. However, these kinds of dialogs are primarily text-based for both the questioner and the answering agent, without any interaction across image and text inputs. In the fashion domain, there have been some early works for the multi-turn retrieval task. Guo \etal \cite{guo2018dialog} proposed a model that uses convolutional neural networks (CNNs) for encoding images and text, followed by a recurrent neural network for aggregating sequences. Then, a $k$-Nearest Neighbor search is performed across sampled images to get the closest match. The entire model is trained end-to-end using reinforcement learning (RL). Inspired by this, Zhang \etal \cite{zhang2019text,zhang2022text} proposed two approaches for enhancing the text-image feature fusion by adding constraints, and using offline interactive recommendation. Recently, Yuan \etal \cite{yuan2021conversational} released the first multi-turn fashion image retrieval dataset, based on the original single turn FashionIQ \cite{Wu_2021_CVPR}. They also proposed a state-of-the-art model, which we directly compare with our approach.

\noindent\textbf{Memory networks for vision and language} - Memory networks have been widely used for a number of natural language processing and computer vision applications. Some works \cite{weston2014memory, sukhbaatar2015end, fan2019heterogeneous, Datta_2022_CVPR} utilize it for Sentence Video Questions and Answering (QA) task. Another popular application is video object segmentation \cite{cheng2022xmem, xie2021efficient, seong2020kernelized, lu2020video}. Recently, there have been some works on including memory in transformer architectures \cite{ji2022lamemo, sandler2022fine, wu2020memformer, https://doi.org/10.48550/arxiv.2109.00301}. However, these approaches design their memory to be used only for specific tasks, and hence cannot be directly compared with ours. In this work, we propose a memory network based method for the multi-turn fashion image retrieval task.

\section{FashionNTM}

The multi-turn feedback-based image retrieval task can be viewed as a series of information exchange transactions. We define a transaction as one session of query context comprising a query image and the associated feedback text. Notationally, an $N$-turn transaction is represented as $T = [(I_1, U_1), (I_2, U_2),\cdots, (I_N, U_N)]$, where $I_n$ and $U_n$ correspond to the query image and feedback utterance respectively, at turn $n \in [1,N]$. Given such a transaction, the aim of a multi-turn model is to iteratively retrieve the final desired target image $I_{tar}(N)$ by ranking candidates in the fashion image database based on a matching score. 

The overall pipeline of our approach, called FashionNTM, is illustrated in Figure \ref{fig: overall_arch}. We start with a single-turn feature extraction module to encode the multi-modal image and text inputs of each turn $n$ in a multi-turn transaction. It comprises two parallel blocks -- (i) a query feature extractor (QE), for processing $I_n$ and $U_n$ to generate the joint query representation $f_{qry}(n)$, and (ii) a target feature extractor (TE), for encoding all the images in the database into their corresponding index features. For the ground-truth target image $I_{tar}(n)$, we call these features $f_{tar}(n)$, while for every other sample $I_{neg}(n)$ in the batch, we name it $f_{neg}(n)$. The query feature $f_{qry}(n)$ is fed turn-wise to the Cascaded Memory Neural Turing Machine (CM-NTM) block. CM-NTM first computes several derived features from the original query feature. Subsequently, the original query feature and each of the derived features interact with their own controllers, read/write heads, and sequentially update the memories in a cascaded manner, \ie output of one memory goes as input to the next. Ultimately, we get the enhanced feature $f'_{qry}(n)$ as the final output, which is then compared with the target feature $f_{tar}(n)$ using a similarity score-based batch loss function. This loss treats $f_{tar}(n)$ as a positive sample and every other feature in the batch, $f_{neg}(n)$, as a negative sample. We use cosine-similarity \cite{vo2019composing, Lee_2021_CVPR, Goenka_2022_CVPR} between feature vectors as the matching score. During inference, the model receives a sequence of multi-modal turn-wise inputs and computes the final modified feature $f'_{qry}(N)$. This is compared with feature $f_{cand}$ derived from different candidate images in the database, and the closest match is retrieved as output. In the following sections, we describe the key components of our framework.

\noindent\textbf{Single-turn feature extractor} - We utilize the state-of-the-art single-turn image retrieval model, FashionVLP \cite{Goenka_2022_CVPR}, for extracting multi-modal text and image features. FashionVLP extracts image embeddings at multiple levels of granularity and incorporates a vision-language pre-trained transformer for fusing these encodings with text feedback to obtain multi-modal query features. It adopts a convolutional neural network (CNN) architecture with contextual attention for fusing fashion-contextual image features of target images. Specifically, for each turn of a multi-turn transaction, we obtain query features $f_{qry}$ and target features $f_{tar}$. An $N$-turn retrieval transaction is of the form $T_{retr} = [f_{qry}(1), f_{qry}(2),\cdots,f_{qry}(N)]$, with the target feature representation given by $f_{tar}(N)$.

\begin{figure}[t]
    \centering
  \includegraphics[width=\linewidth]{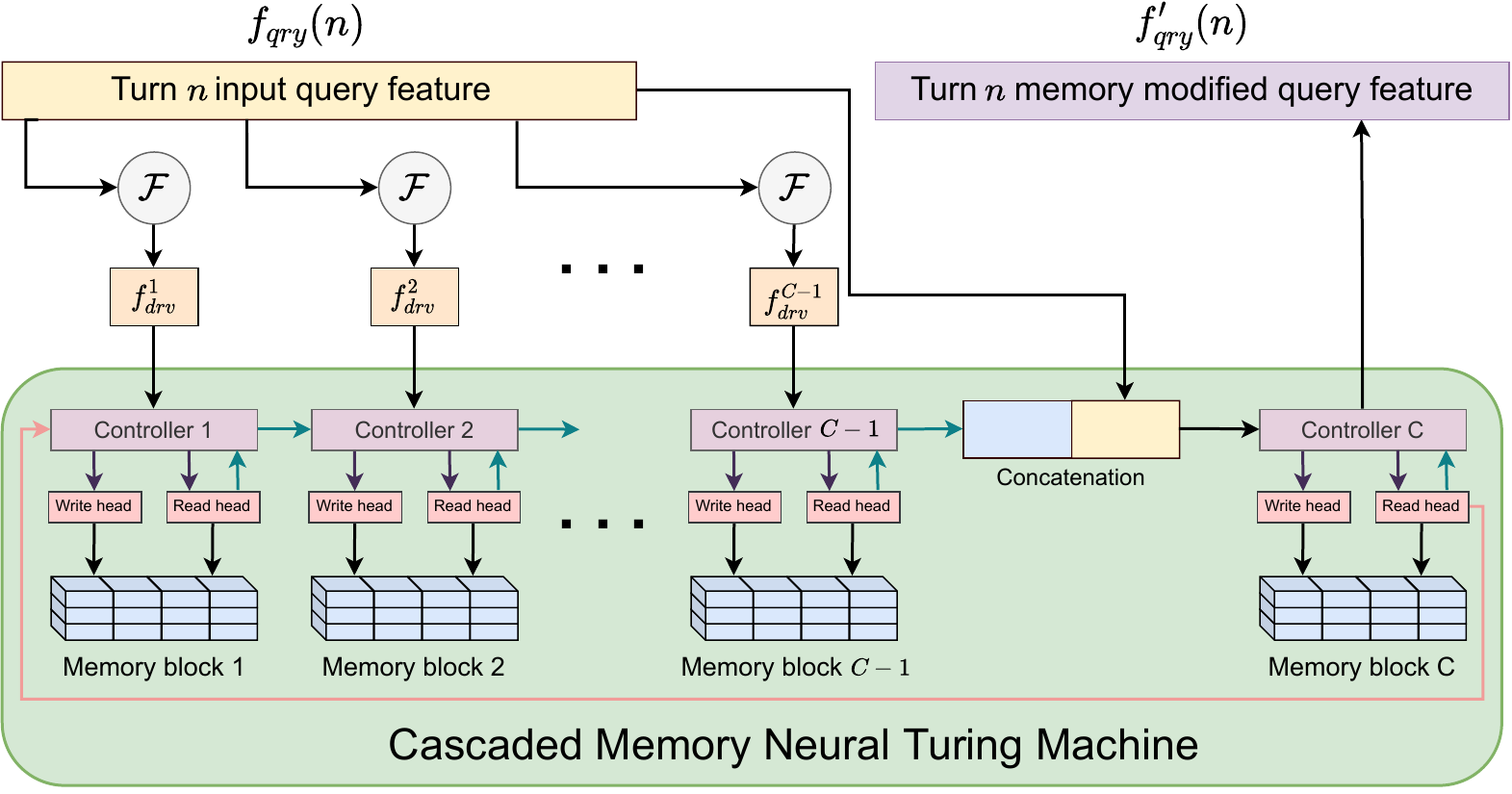}
  \caption{Cascaded Memory Neural Turing Machine (CM-NTM). It consists of $C$ controllers; one for the query feature and $C-1$ for the derived features. Each controller interacts with its own memory using its own read and write heads. The controllers are linked with each other, forming a cascaded chain. The modified features from the last memory are output as the final embedding.}
  \label{fig: ntm_arch}
  \vspace{-15pt}
\end{figure}

\subsection{Cascaded Memory Neural Turing Machine}
To learn relationships across transactions, we propose a novel Cascaded Memory Neural Turing Machine (CM-NTM) module. CM-NTM allows multiple inputs to interact with their own memories using individual read and write heads to learn multiple complex relationships in the input data. We achieve this by deriving several features from the original $n$-th turn query feature $f_{qry}(n)$ using a projection function $\mathcal{F}$, which comprises a fully-connected (FC) layer with batch normalization. Specifically, for a $C$-stage cascaded CM-NTM, we obtain the derived feature $f^c_{drv} = \mathtt{BatchNorm}(\mathtt{FC_c}(f_{qry}(n)))$, where $c \in [1,C-1]$, and $\mathtt{FC_c}(\cdot)$ is the FC layer for the $c$-th stage. Having obtained $C$ inputs comprising the original query and $C-1$ derived features, we pass them sequentially to our memory network. \Cref{fig: ntm_arch} presents our memory network architecture. It consists of three main components\footnote{We mainly discuss our novel changes to the NTM design. For details regarding the vanilla architecture, please refer to the original paper \cite{graves2014neural}.}: Controller, Read/Write heads, and Memory blocks.

\noindent\textbf{Controller} - The vanilla NTM has a single controller which takes the query feature at turn $n$, along with the previous turn's read vector, and emits an intermediate controller output. This is used by the read and write heads to compute the current turn's attention weights, which are then used to update the memory. In our work, we introduce $C$ different controllers -- one for each of our inputs. Each controller, $c \in [1, C]$ can therefore interactively update its memory via individual read and write heads, allowing it to learn multiple complex relationships in each turn.

\noindent\textbf{Read/Write heads} - The controller output is fed to these heads. The write head learns to generate erase and add parameters, which are used to update the current memory. Similarly, the read head generates an attention weight vector, which is used to obtain a weighted sum over the memory locations to get the read vector $\mathtt{r}_{out}(n)$. In our work, we have separate read and write heads for each memory. 

\noindent\textbf{Memory block} - This is represented as a $2$-D matrix of the form $N \times M$, where $N$ corresponds to memory locations and $M$ to the vector size at each location. The output of each memory block is a fused representation of the controller output and the read vector. In our cascaded multi-memory setup, the controllers are linked in a chain, such that the memory-modified features are sequentially propagated. Specifically, for controller $c \in [1,C]$ at turn $n$, the input is given by $\mathtt{input}^c(n) = [\mathtt{r}_{out}^{c-1}(n);f^c_{drv};\mathtt{r}_{out}^{c}(n-1)]$, where $\mathtt{r}_{out}^{0}(n) = \mathtt{r}_{out}^{C}(n-1)$, $f^C_{drv} = f_{qry}(n)$, and $;$ represents concatenation. The final output, $f'_{qry}(n)$, is the fused representation of the last controller output $\mathtt{output}_{ctrl}^{C}$ and final read vector $\mathtt{r}_{out}^{C}$. We experiment with a different number of memories $C$ and memory sizes in \Cref{sect: abl_study}.
\begin{table*}[!b]
\renewcommand\thetable{2}
\scriptsize
\centering
\caption{Quantitative results on Multi-turn FashionIQ \cite{yuan2021conversational}. We compare with multiple single-turn and multi-turn baselines, and state-of-the-art works \cite{guo2018dialog, yuan2021conversational}. Results show the superior performance of our proposed approach on the popularly used recall rate evaluation metric.}
\label{tab:sota}
\begin{tabular}{lccccccccccccc}\toprule
 && \multicolumn{2}{c}{\textbf{Dress}} && \multicolumn{2}{c}{\textbf{Toptee}} && \multicolumn{2}{c}{\textbf{Shirt}} && \multicolumn{3}{c}{\textbf{Overall}}\\
\cline{3-4} \cline{6-7} \cline{9-10} \cline{12-14}
\textbf{Model} && \textbf{R@$\mathbf{5}$} & \textbf{R@$\mathbf{8}$ }&& \textbf{R@$\mathbf{5}$} & \textbf{R@$\mathbf{8}$} && \textbf{R@$\mathbf{5}$} & \textbf{R@$\mathbf{8}$} && \textbf{R@$\mathbf{5}$} & \textbf{R@$\mathbf{8}$} & \textbf{Mean}\\
\hline
\multicolumn{14}{c}{\textit{Single-turn}} \\
\hline
ST + avg(all turns) && $25.6$ & $32.5$ && $32.1$ & $38.1$ && $27.0$ & $32.3$ && $28.2$ & $34.3$ & $31.3$ \\
ST + cat(all captions) && $30.2$ & $36.3$ && $36.0$ & $44.1$ && $35.3$ & $42.5$ && $33.8$ & $41.0$ & $37.4$ \\
\hline
\multicolumn{14}{c}{\textit{Multi-turn}} \\
\hline
Dialog Manager \cite{guo2018dialog} && $12.7$ & $16.7$ && $11.6$ & $15.8$ && $13.9$ & $17.7$ && $13.1$ & $15.2$ & $14.2$ \\
CFIR \cite{yuan2021conversational} && $29.8$ & $33.5$ && $29.4$ & $33.6$ && $30.5$ & $34.1$ && $30.3$ & $33.4$ & $31.9$ \\
ST + EWMA (ours)  && $42.0$ & $48.4$ && $43.8$ & $\mathbf{50.9}$ && $36.9$ & $44.2$ && $40.9$ & $47.8$ & $44.4$ \\
ST + LSTM (ours) && $47.8$ & $52.5$ && $44.4$ & $50.5$ && $41.6$ & $47.6$ && $44.6$ & $50.2$ & $47.4$ \\
\hline
FashionNTM (ours) && $\mathbf{48.3}$ & $\mathbf{52.8}$ && $\mathbf{45.1}$ & $49.8$ && $\mathbf{43.8}$ & $\mathbf{48.8}$ && $\mathbf{45.7}$ & $\mathbf{50.4}$ & $\mathbf{48.1}$ \\
\bottomrule
\vspace{-15pt}
\end{tabular}
\end{table*}

\subsection{Loss function}
We adopt a batch cross-entropy loss \cite{Goenka_2022_CVPR}, where each entry in a batch acts as a negative sample for all other entries. In the multi-turn setting, we compute the loss function turn-wise. Given a batch size $B$, the loss between predicted feature $\mathbf{x}_n=[^{1}f'_{qry}(n), ^{2}f'_{qry}(n), \hdots, ^{B}f'_{qry}(n)]$, and ground-truth $\mathbf{y}_n=[^{1}f_{tar}(n), ^{2}f_{tar}(n), \hdots, ^{B}f_{tar}(n)]$, at turn $n$ is of the form
\begin{equation*}
    \mathcal{L}(\mathbf{x}_n,\mathbf{y}_n) = \frac{1}{B}\sum_{i=1}^B-\log \frac{e^{\kappa (^{i}f'_{qry}(n), ^{i}f_{tar}(n))}}{\sum_{j=1}^Be^{\kappa (^{i}f'_{qry}(n), ^{j}f_{tar}(n))}}
\end{equation*}
\begin{table}[t]
\renewcommand\thetable{1}
\scriptsize
\centering
\caption{Statistics of the datasets used in this work. The top $2$ rows are for the recently proposed Multi-turn FashionIQ dataset \cite{yuan2021conversational}. The bottom two rows are for the Multi-turn version of the Shoes \cite{berg2010automatic} dataset, which we created as part of our work.}
\label{tab:dataset_stat}
\setlength{\tabcolsep}{5pt}
\begin{tabular}{l|rrrrr}\toprule
 &  & \multicolumn{3}{c}{\textbf{Number of transactions with}} & \\
 \cline{3-5}
\multirow{2}{*}{\textbf{Dataset}} & \multirow{2}{*}{\textbf{Split}} & \multirow{2}{*}{\textbf{$2$-turns}} & \multirow{2}{*}{\textbf{$3$-turns}} & \multirow{2}{*}{\textbf{$4$-turns}} & \textbf{Total}\\
&&&&& \textbf{images} \\
\hline
\multirow{2}{*}{Multi-turn FashionIQ \cite{yuan2021conversational}} & Train & $6,897$ & $1,733$ & $475$ & $10,438$ \\
 & Test & $1,752$ & $483$ & $165$ & $6,274$ \\
\hline
\multirow{2}{*}{Multi-turn Shoes (ours)} & Train & $1,659$ & $1,036$ & $982$ & $11,030$ \\
 & Test & $296$ & $96$ & $28$ & $4,631$ \\
\bottomrule
\end{tabular}
\vspace{-5pt}
\end{table}

where $\kappa$ is the cosine similarity metric \cite{Goenka_2022_CVPR}. In this way, each $^{j}f_{tar}(n)$ in the batch, $\forall j \in [1,B] \text{ and } j\neq i$, serves as a negative sample $f_{neg}(n)$ for a given $^{i}f'_{qry}(n)$. The turn-wise retrieval loss function 
is represented as $\mathcal{L}^n_{retr}=\mathcal{L}(\mathbf{x}_n,\mathbf{y}_n)$, with the overall loss function for our proposed multi-turn model given by $\mathcal{L} = \frac{1}{N}\sum_{n=1}^N \mathcal{L}^n_{retr}$.

\section{Experimental Evaluation}
\begin{table*}[!b]
\renewcommand\thetable{4}
\scriptsize
\centering
\caption{Importance of aggregating historical data using memory-based vs non-memory approach. In the first row, we show the result of a model using only the final turn information of a multi-turn transaction. This assumes the groundtruth retrieval for all previous turns, and therefore provides the upper-bound on single-turn performance for final retrieval. Subsequently, we include additional information from the history, and compare performance across models with and without memory. As seen from the non-memory case, the performance depreciates a lot ($\approx 64.9\%$ difference compared to the final turn's performance), as we try to aggregate longer historical information. In contrast, for the memory network model, it can be seen that the performance is fairly consistent ($\approx 0.5\%$ difference) across the turn length.}
\label{tab:turns_comp}
\begin{tabular}{l|c|ccccccccccccc|cc}\toprule
\multirow{2}{*}{{\textbf{Memory usage}}} & \multirow{2}{*}{{\textbf{Turn configuration}}} && \multicolumn{2}{c}{\textbf{Dress}} && \multicolumn{2}{c}{\textbf{Toptee}} && \multicolumn{2}{c}{\textbf{Shirt}} && \multicolumn{3}{c|}{\textbf{Overall}} & \textbf{Difference}\\
\cline{4-5} \cline{7-8} \cline{10-11} \cline{13-15}
 &&& \textbf{R@$\mathbf{5}$} & \textbf{R@$\mathbf{8}$} &&\textbf{ R@$\mathbf{5}$} &\textbf{ R@$\mathbf{8}$} && \textbf{R@$\mathbf{5}$} &\textbf{ R@$\mathbf{8}$} && \textbf{R@$\mathbf{5}$} & \textbf{R@$\mathbf{8}$} & \textbf{Mean} & \textbf{from final turn}\\
\hline
\multicolumn{1}{c|}{-} & only final turn && $77.9$ & $77.9$ && $84.0$ & $84.0$ && $74.1$ & $77.8$ && $78.7$ & $79.9$ & $79.3$ & -\\
\hline
\multicolumn{16}{c}{\textit{Experiments with data aggregated across multiple-turns}} \\
\hline
\multirow{3}{*}{\parbox{1.6cm}{Only single turn (w/o memory)}} & last two turns && $51.3$ & $58.4$ && $56.0$ & $76.0$ && $44.4$ & $48.1$ && $50.6$ & $60.9$ & $55.8$ & \color{red}$-29.6\%$\\
& last three turns && $33.6$ & $43.4$ && $32.0$ & $44.0$ && $29.6$ & $40.7$ && $31.8$ & $42.7$ & $37.3$ & \color{red}$-53.0\%$\\
& all turns && $18.6$ & $28.3$ && $32.0$ & $32.0$ && $25.9$ & $29.6$ && $25.5$ & $30.0$ & $27.8$ & \color{red}$-64.9\%$\\
\hline
\multirow{3}{*}{\parbox{1.6cm}{FashionNTM (with memory)}} & last two turns && $76.1$ & $77.9$ && $84.0$ & $84.0$ && $77.8$ & $77.8$ && $79.3$ & $79.9$ & $79.6$ & \color{green}$+0.4\%$\\
& last three turns && $77.0$ & $77.9$ && $84.0$ & $84.0$ && $77.8$ & $77.8$ && $79.6$ & $79.9$ & $79.8$ & \color{green}$+0.6\%$\\
& all turns && $76.1$ & $77.9$ && $84.0$ & $84.0$ && $77.8$ & $77.8$ && $79.3$ & $79.9$ & $79.6$ & \color{green}$+0.4\%$\\
\bottomrule
\end{tabular}
\vspace{-10pt}
\end{table*}

\subsection{Datasets}
\noindent\textbf{Multi-turn FashionIQ} \cite{yuan2021conversational}: To the best of our knowledge, this is the only existing fashion dataset with multi-turn sessions, where each turn is derived from the original single-turn FashionIQ~\cite{Wu_2021_CVPR} dataset. It comprises of $11,505$ sessions across three clothing types -- dress, top-tee, and shirt. The dataset is split into transactions of $2$-turns, $3$-turns, and $4$-turns. In each turn, the data is represented as a pair ($I_n$, $U_n$), where $I_n$ and $U_n$ correspond to the query image and the feedback text for turn $n$.

\noindent\textbf{Multi-turn Shoes}: The original Shoes dataset \cite{berg2010automatic} contains images of $10$ categories of women's shoes obtained from the web along with automatic labeling of attributes. Guo \etal \cite{guo2018dialog} provide additional natural language descriptions of the images to make them suitable for \textit{single-turn} feedback-based image retrieval. This resulted in about $10$k training pairs and $4.6$k test queries. In this work, we create a multi-turn extension of this dataset to further research in this domain. Like the approach described in \cite{yuan2021conversational}, we concatenated several single-turn transactions by matching the target image of one session to the query of another. To maintain consistency with Multi-turn FashionIQ, we also developed transactions of $2$-turns, $3$-turns, and $4$-turns. \Cref{tab:dataset_stat} provides the statistics of both the datasets.

\subsection{Implementation Details} \label{sect: impl_detail}
\noindent\textbf{Single-turn feature extractor pre-training} - We use the recently proposed single turn (ST) fashion image retrieval model, FashionVLP \cite{Goenka_2022_CVPR}, to extract the query and target features for both datasets. To ensure a fair comparison with other algorithms, we re-train FashionVLP only on those single-turn queries that are part of the multi-turn dataset. The implementation details and hyperparameters are similar to those mentioned in Goenka \etal \cite{Goenka_2022_CVPR}.

\noindent\textbf{CM-NTM training} - We build our CM-NTM model using the open-source implementation \cite{clemkoantm2019} of NTM \cite{graves2014neural}. We implement a separate controller for each memory in our cascaded design using Long Short Term Memory (LSTM) networks~\cite{hochreiter1997long}. The read and write heads are composed of multi-layer perceptrons (MLPs) and attention networks. We use $C=4$ for Shoes, and $C=8$ memory stages for FashionIQ, as it is a larger dataset. To ensure equal turn lengths for batch training, we pad the short turn transactions by repeating the last transaction similar to Yuan \etal \cite{yuan2021conversational}. We train our models for $100$ epochs with a batch size of $80$ and a learning rate of $1$e-$4$. We used the PyTorch framework with Adam \cite{kingma2014adam} optimizer for training. 

\subsection{Baselines and Previous State-of-the-art}

We compare the performance of our model with six other single-turn and multi-turn approaches.

\noindent\textbf{Single-turn methods} - In these methods, we aggregate the multi-turn query data into a \emph{single} feature without iterating over them turn-wise. The first baseline is $\mathtt{ST+avg(all~turns)}$, where we take the mean of all the $N$ query features in a transaction to get a single mean query $\bar{f}_{qry}(N)$. This is compared with the candidate target features $f_{tar}(N)$. Next, we have $\mathtt{ST+cat(all~captions)}$, where all the captions of a multi-turn transaction are concatenated into one long caption, along with initial reference image, to get query features $\hat{f}_{qry}(N)$. This is compared with target features $f_{tar}(N)$ to retrieve the final images.

\noindent\textbf{Multi-turn methods} - In these methods, the multi-turn data is fed turn-wise to the model. The first method is $\mathtt{Dialog~Manager}$ \cite{guo2018dialog}, which employs reinforcement learning (RL) framework to learn relationships between turns. Next, we have the previous state-of-the-art Conversational Fashion Image Retrieval $\mathtt{CFIR}$ method \cite{yuan2021conversational}, which encodes \textit{only} the text using a transformer network \cite{NIPS2017_3f5ee243}, and uses a simple Gated Recurrent Unit (GRU) layer for multi-turn image retrieval. The third baseline is $\mathtt{ST+EWMA}$, where we use exponential weighted moving average~\cite{ewma} to \textit{heuristically} aggregate past turns, with more weights given to recent history. Finally, we have the $\mathtt{ST+LSTM}$ model that uses a single-layer LSTM~\cite{hochreiter1997long} with a hidden size of 100 for aggregating past information.

\subsection{Results}

\noindent\textbf{Evaluation Metrics} - Following ~\cite{Goenka_2022_CVPR, yuan2021conversational, guo2018dialog}, we evaluate models using the standard top-$\mathtt{K}$ recall (\ie R@$\mathtt{K}$) for image retrieval. Overall performance are compared specifically on the average of R@$5$ and R@$8$.

\noindent\textbf{Quantitative results on multi-turn datasets} - In \Cref{tab:sota}, we compare the results of different methods on the multi-turn FashionIQ dataset. We observe that the multi-turn baselines generally perform better than single-turn methods. This is expected as aggregating data by na\"ively averaging/concatenating loses feedback content and turn-order information and hence is likely to miss out on important cues. Our memory-based approach outperforms all the other multi-turn baselines by a large margin. This shows that the memory network can store and retrieve useful information to and from the memory between intermediate turns, which allows it to keep track of past information better than other networks that do not use explicit memory. The $50.5\%$ performance gain over the previous state-of-the-art highlights our model's capability to learn meaningful representations over multiple turns of conversational feedback.

In \Cref{tab:shoes_res}, we provide a similar comparison for the Multi-turn Shoes dataset. Consistent with results for Multi-turn FashionIQ, the multi-turn baselines perform better than all the single-turn ones, and our memory network-based approach performs the best with a relative improvement of $12.6\%$. The difference in performance across different models is more pronounced in these results than in \Cref{tab:sota}. This is possibly because the annotations are cleaner and more consistent in this dataset as compared to Multi-turn FashionIQ \cite{yuan2021conversational}. For instance, multiple images in the FashionIQ dataset can match a particular query, but only one of them is labeled as the ground-truth. 

\begin{table}[t]
\renewcommand\thetable{3}
\scriptsize
\centering
\caption{Comparison with existing single-turn and multi-turn models on the multi-turn version of the Shoes \cite{berg2010automatic}  dataset. We compare with multiple single-turn, and multi-turn baselines. Comparative analysis shows the superior performance of our proposed approach using the popular recall rate evaluation metric.}
\label{tab:shoes_res}
\begin{tabular}{lcccccccc}\toprule
\textbf{Model} & \multicolumn{2}{c}{\textbf{R@$\mathbf{5}$}} & \multicolumn{2}{c}{\textbf{R@$\mathbf{8}$}} & \multicolumn{2}{c}{\textbf{Mean}}\\
\hline
\multicolumn{7}{c}{\textit{Single-turn }} \\
\hline
ST + avg(all turns) & \multicolumn{2}{c}{$12.4$} & \multicolumn{2}{c}{$17.6$} & \multicolumn{2}{c}{$15.0$} \\
ST + cat(all captions) & \multicolumn{2}{c}{$10.7$} & \multicolumn{2}{c}{$13.6$} & \multicolumn{2}{c}{$12.2$} \\
\hline
\multicolumn{7}{c}{\textit{Multi-turn }} \\
\hline
ST + EWMA (ours) & \multicolumn{2}{c}{$18.3$} & \multicolumn{2}{c}{$23.8$} & \multicolumn{2}{c}{$21.1$} \\
ST + LSTM (ours)& \multicolumn{2}{c}{$23.3$} & \multicolumn{2}{c}{$32.1$} & \multicolumn{2}{c}{$27.7$} \\
\hline
FashionNTM (ours)& \multicolumn{2}{c}{$\mathbf{26.7}$} & \multicolumn{2}{c}{$\mathbf{35.7}$} & \multicolumn{2}{c}{$\mathbf{31.2}$} \\
\bottomrule
\vspace{-20pt}
\end{tabular}
\end{table}
\begin{figure*}[t]
\centering
\subfloat[Multi-turn FashionIQ. From left to right, the images belong to the Dress, Shirt, and Toptee categories respectively.]{%
  \includegraphics[width=\linewidth]{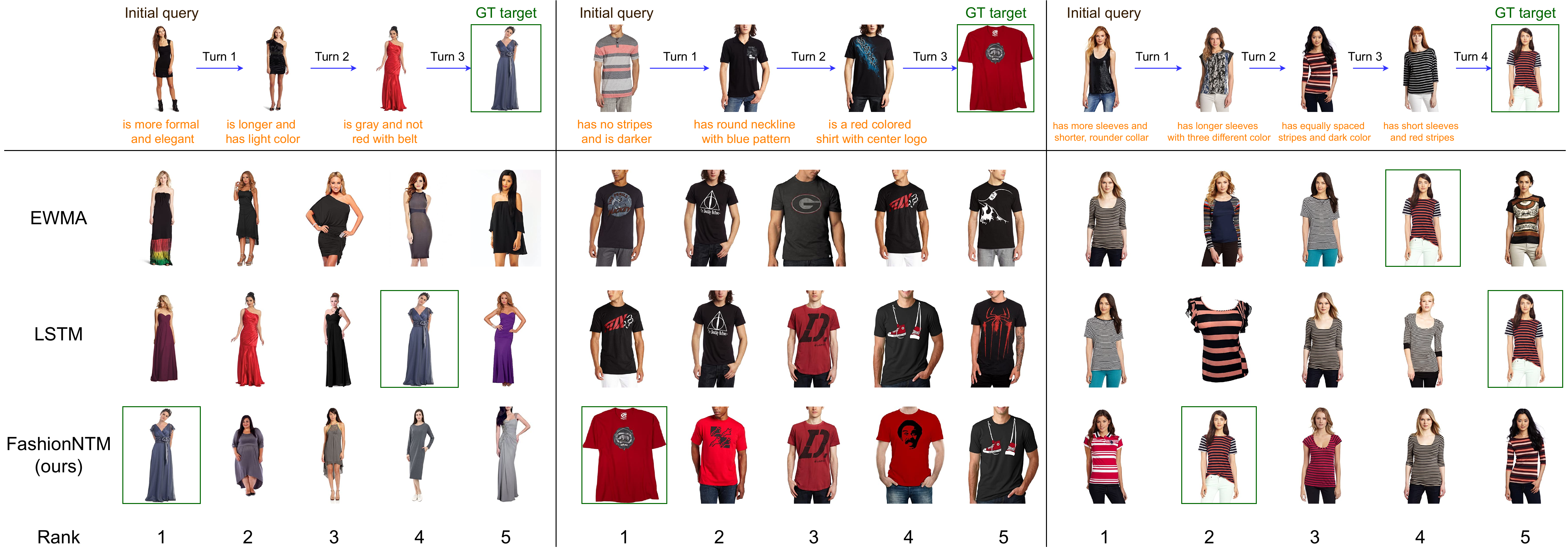}
  \label{fig: topk_qual_fiq}%
}
\\
\subfloat[Multi-turn Shoes. From left to right, we have three different samples of shoes from the dataset.]{%
  \includegraphics[width=\linewidth]{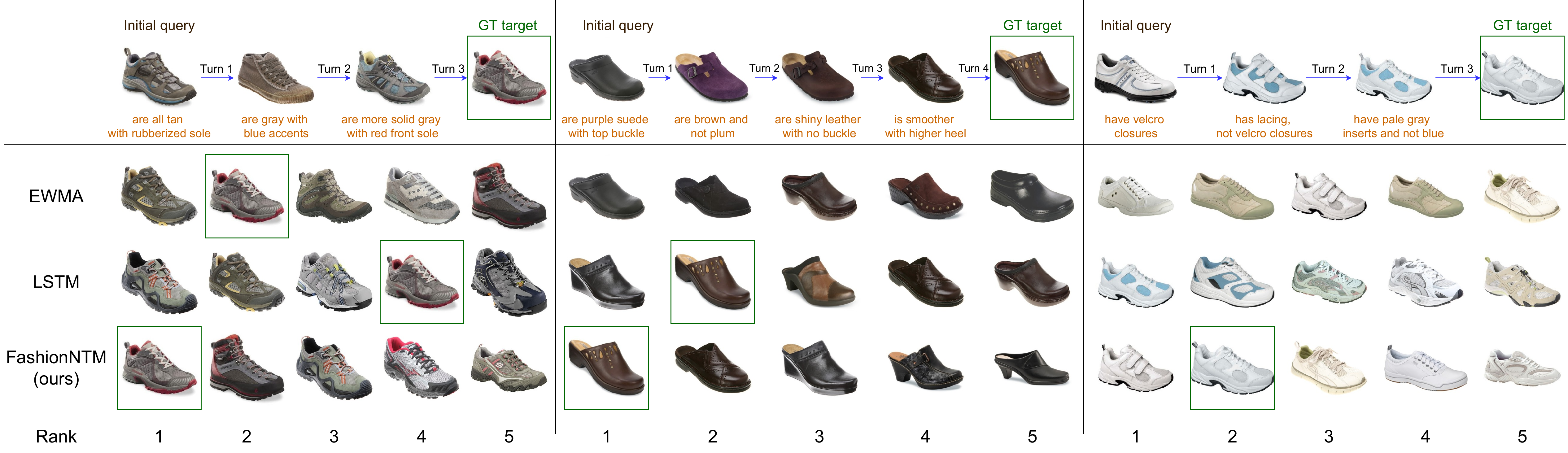}
  \label{fig: topk_qual_shoes}%
}
\caption{Top-5 final image retrievals on the evaluated multi-turn datasets. The top row illustrates $3$ sets of multi-turn query session. We consider three different multi-turn models - $\mathtt{EWMA}$, $\mathtt{LSTM}$, and our proposed $\mathtt{FashionNTM}$. As seen from the retrievals, our proposed model correctly predicts the target image in all $3$ cases for the FashionIQ dataset, and in $2$ out of $3$ cases for the Shoes dataset.}
\end{figure*}

An important property of a good multi-turn system is that performance should be robust to the length of the historical information (number of past turns). For example, even for a large number of previous turns considered, the model should efficiently retain desirable details, while filtering out unnecessary information. In Table \ref{tab:turns_comp}, we analyze this property for models with and without memory. A single-turn model that assumes ground-truth information for all past turns, and evaluated only on the final-turn, is taken as reference. This is expected to be the upper-bound on the performance in a single-turn setting, as perfect information about the history is guaranteed. We vary the number of past turns included in the input transaction history (versus treated as ground-truth) for the multi-turn models (with or without memory) in order to evaluate their effectiveness at capturing and utilizing past information. As seen from the table, for a model without memory, the performance depreciates significantly with each additional turn from the past treated as input rather than ground-truth. As we go further back in history, the performance consistently reduces. However, for our multi-turn model with memory, the performance does not change appreciably as the number of turns change. This shows that our proposed approach can successfully retain/filter out past data based on their relevance, across various lengths of history.

An interesting observation is that having only the final turn with memory does not yield a good result. This is possibly due to the initialization method of the memory network, which is random. Hence, in absence of a history (only single turn case), the only past features to be aggregated are the random initialization features.

\noindent\textbf{Qualitative results} - In addition to the quantitative experiments described above, we also present some qualitative final image retrieval results of our evaluated models on both the multi-turn datasets. The first set of results are shown in Figures \ref{fig: topk_qual_fiq} and \ref{fig: topk_qual_shoes} for the Multi-turn FashionIQ \cite{yuan2021conversational} and the Multi-turn Shoes datasets, respectively\footnote{Please refer to the supplementary material for additional results, along with the differences in annotation quality between the two datasets.}. We compare the top-$5$ predicted results from our proposed model with two other multi-turn baselines, $\mathtt{ST+EWMA}$ and $\mathtt{ST+LSTM}$. As seen in the figures, our approach can correctly predict the desired target image for both datasets with higher ranks as compared to other baselines. More specifically, we see that in \Cref{fig: topk_qual_fiq}, even though all the three multi-turn models can infer the general sense of desired attributes, such as ``is longer'' in the left block and ``has short sleeves'' in the right, only our model can capture complex and detailed properties, \eg, ``gray and not red'' in the left block, and ``red color with center logo'' in the middle block. Furthermore, our model can retrieve multiple desirable products, as seen by the first four images in the middle block, and four out of five images in the right block. Similar results are observed for Shoes in \Cref{fig: topk_qual_shoes}, where our model correctly predicts the desired target as rank-$1$ in two out of three examples, whereas the other models fail to retrieve meaningful results.

\subsection{Model Analysis in Interactive Settings}
In addition to the results above for the static dataset, we also performed some interactive experiments to evaluate whether our model can adapt to real-world dynamic use cases beyond the trained datasets.

\begin{figure}[t]
    \centering
  \includegraphics[width=\linewidth]{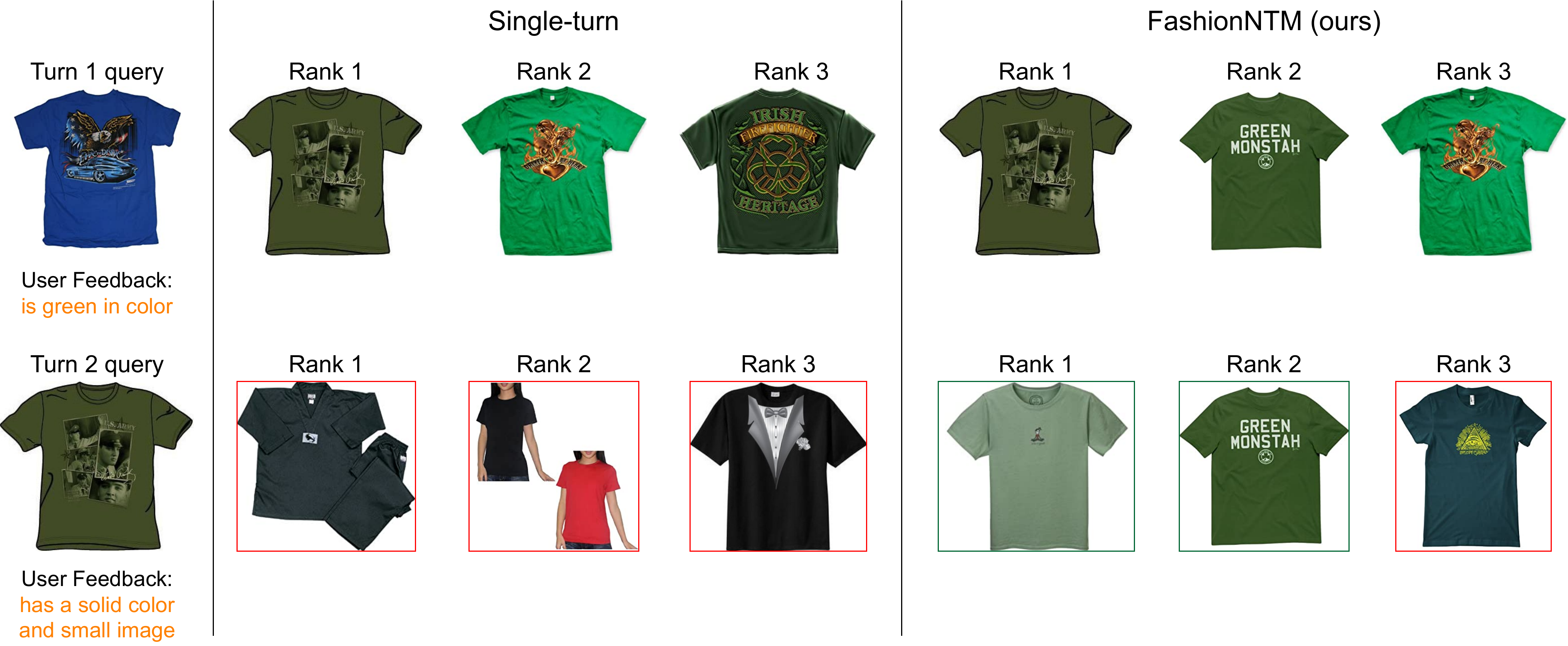}
  \caption{Memory retention capability of our proposed approach. Given an initial query image, we take two interactive user feedbacks in turns. On the left side, we have the single-turn model which only learns to retrieve an image using a single dialog exchange. As a result, none of the retrieved images in turn $2$ are ``green in color", which was desired in turn $1$. In contrast, our proposed approach on the right can learn to retain data from both the turns, and therefore retrieves desirable product in $2$ out of $3$ cases.}
  \label{fig: memret_qual}
  \vspace{-15pt}
\end{figure}

\noindent\textbf{Memory retention} - In this experiment, we demonstrate the memory retention capability of our proposed model by comparing it with a single-turn network, which does not retain historical information. We start with an initial query image retrieved via a single-turn model. Subsequently, we take user input for the next two turns to retrieve newer sets of images. As seen in \Cref{fig: memret_qual}, for the single-turn model (left side), none of the retrieved images in turn $2$ are \textit{green} in color. This is because the ``green in color'' attribute was a desired property in turn $1$, which the model without memory could not recover. In contrast, for our memory network approach (right side) both the top-$2$ retrieved images for turn $2$ are ``green in color'' in addition to having ``a solid color and small image''. Thus, our model can learn to retain information from previous turns.

\begin{figure}[t]
    \centering
  \includegraphics[width=\linewidth]{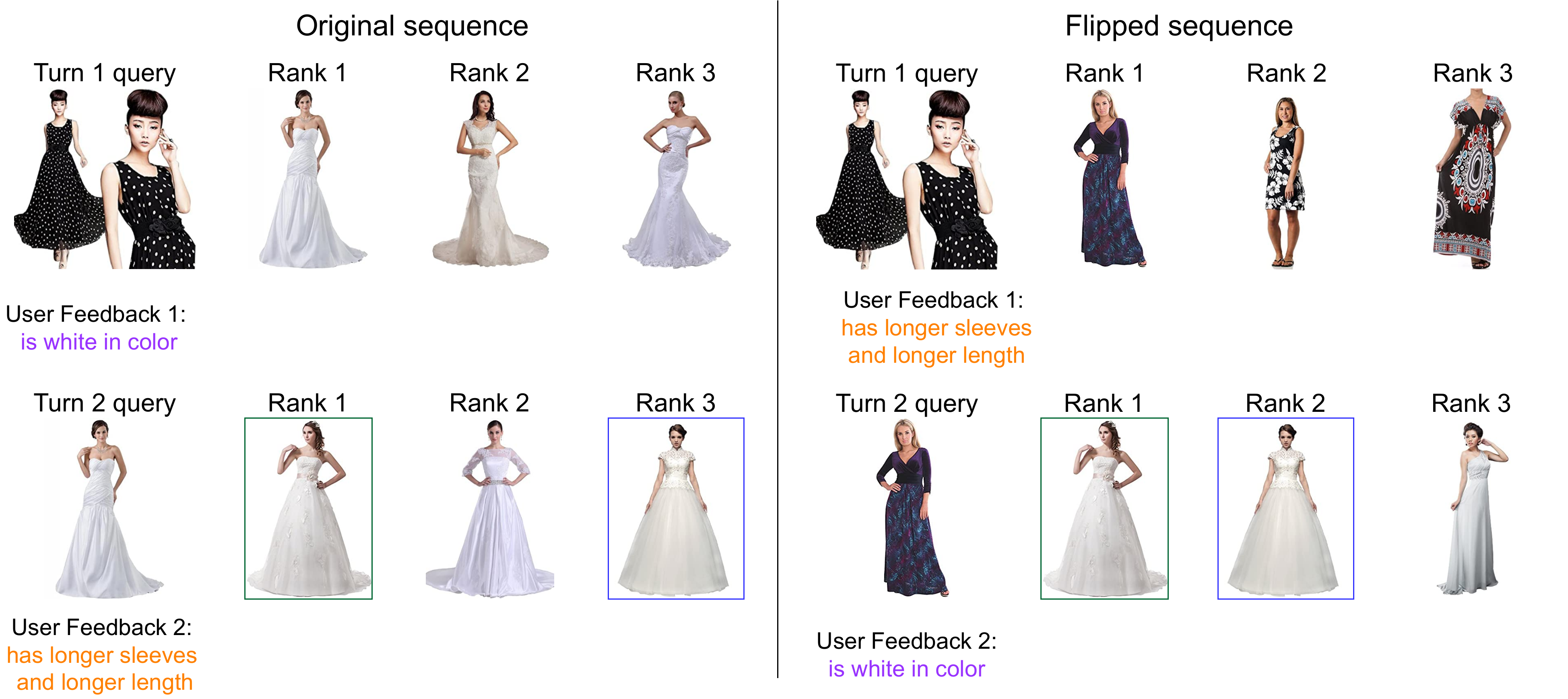}
  \caption{Turn order independence feature of a memory network. In this experiment, we start with an initial query image, and take two \textit{non-contradictory} user feedbacks. In one case (left), we let the model retrieve images based on the original order of the feedbacks, whilst in the other (right), we flip the order of the feedbacks. Our proposed approach adapts to presented input, and retrieves similar looking final results for both the cases, even though the intermediate outputs are quite different.}
  \label{fig: turn_order_qual}
  \vspace{-12pt}
\end{figure}

\noindent\textbf{Agnosticity to turn order} - Ideally, a deployed multi-turn image retrieval system should be independent of the order of feedbacks provided, as long as they are non-contradictory. This is demonstrated in the experiment conducted in \Cref{fig: turn_order_qual}. In the first case, we take two text inputs as feedbacks from a user and present them to the model. In the second case, we reverse the order of the feedbacks. As shown in the figure, for the flipped case, our proposed FashionNTM model retrieves similar looking final products, even though the intermediate retrievals are very different. 
\begin{figure}[!b]
  \centering\includegraphics[width=0.9\linewidth,trim={0.3cm 0 0 0.25cm},clip]{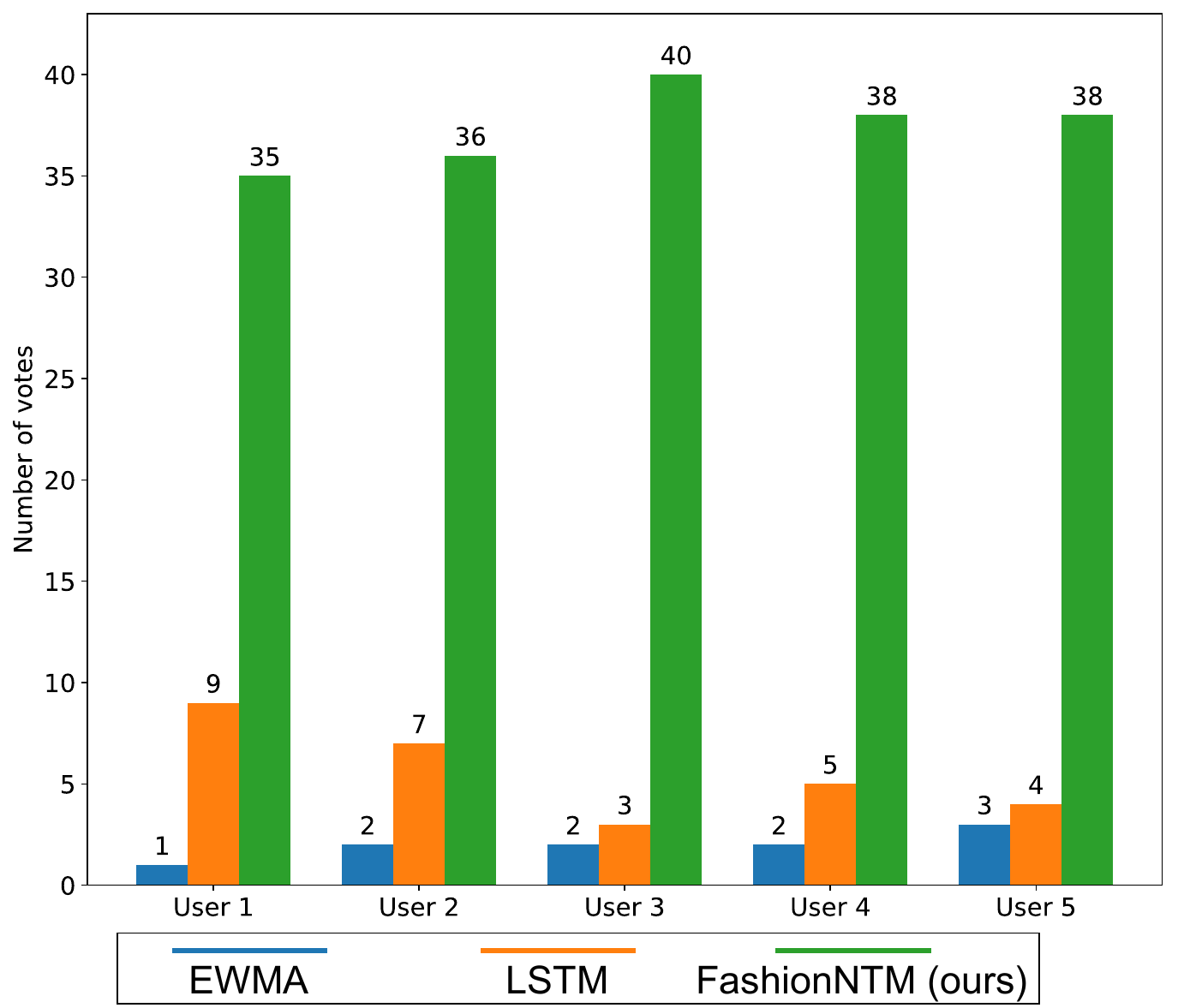}
  \caption{Human preference study of top multi-turn systems.}
  \label{fig: rebuttal_plot}
  \vspace{-15pt}
\end{figure}

It is to be noted that for both the experiments depicted in Figures \ref{fig: memret_qual} and \ref{fig: turn_order_qual}, the feedback is taken from a dynamic user, thereby establishing the interactive capability of our proposed model beyond the training dataset. 

\subsection{User Study}

Fashion image retrieval is inherently a subjective task, where the task success heavily relies on the satisfaction of a customer. Thus, we conducted a small human-preference survey among $5$ participants (not associated with the paper in any way). To each user, we showed the final top-$1$ retrieved image by the $3$-best multi-turn models on the FashionIQ \cite{yuan2021conversational} dataset from \Cref{tab:sota} -- $\mathtt{EWMA}$, $\mathtt{LSTM}$, and our proposed $\mathtt{FashionNTM}$. To ensure consistency, we generated $45$ queries for this study, whose results are shown in \Cref{fig: rebuttal_plot}. The $y$-axis shows the number of preferred retrieval results for each user. Results show that each of the $5$ users preferred images retrieved by the proposed FashionNTM model, on an average $83.1\%$ more as compared to other multi-turn methods. In the supplementary material, we include some examples of the user interface shown to the participants during this study.

\begin{table}[t]
\scriptsize
\centering
\caption{Different number of memories for the proposed approach by fixing the memory size to $4 \times 768$. We select the mean value for comparison and pick the best one.}
\label{tab:num_mem_ablation}
\begin{tabular}{c|c|ccc|c}\toprule
\multirow{2}{*}{\textbf{Model}} & \textbf{Number of}& \multirow{2}{*}{\textbf{R}@$\mathbf{5}$} & \multirow{2}{*}{\textbf{R}@$\mathbf{8}$} & \multirow{2}{*}{\textbf{Mean}} & $\mathbf{\%}$\\
&\textbf{memories (C)}&&&&\textbf{increase}\\
\hline
ST+v-NTM & $1$ & $24.5$ & $30.5$ & $27.5$ & -\\
\hline
\multirow{4}{*}{FashionNTM} &$2$ & $26.4$ & $33.8$ & $30.1$ & $9.5$\\
&$4$ & $\textbf{27.6}$ & $\mathbf{35.5}$ & $\mathbf{31.5}$ & $\textbf{14.5}$\\
&$8$ & $26.9$ & $35.2$ & $31.1$ & $13.1$\\
&$16$ & $26.4$ & $33.8$ & $30.1$ & $9.5$\\
\bottomrule
\end{tabular}
\vspace{-12pt}
\end{table}
\subsection{Ablation Studies} \label{sect: abl_study}
We perform multiple ablation studies to gain insights on how changing different configurations of the memory network affect the overall performance. We perform these ablations on the multi-turn Shoes dataset as it contains more realistic and consistent feedback texts.\footnote{For a similar study on the Multi-turn Fashion-IQ dataset, please refer to the supplementary material.}


\noindent\textbf{Number of memories in CM-NTM} - This experiment involves varying the number of memories $C$ in our cascaded memory architecture. In \Cref{tab:num_mem_ablation}, we observe that the cascaded memory CM-NTM models perform significantly better than vanilla NTM, which has only one memory. We hypothesize that having inputs from multiple turns interacting with the same memory could eventually lead to saturation as we get additional data which could be alleviated if there are multiple memories in the network to recover the past which might help in capturing multiple complex relationships in multi-turn interactions better. Additionally, multiple memory networks can help in learning diverse representations of the input using derived features, which is not possible with a single memory network. As seen in \Cref{tab:num_mem_ablation} for Shoes dataset, the performance increases with the number of memories, peaking at $C=4$, and then gradually decreases as the model starts to overfit.

\begin{figure}[t]
    \centering
  \includegraphics[width=\linewidth,trim={0.5cm 0.5cm 0 0.5cm},clip]{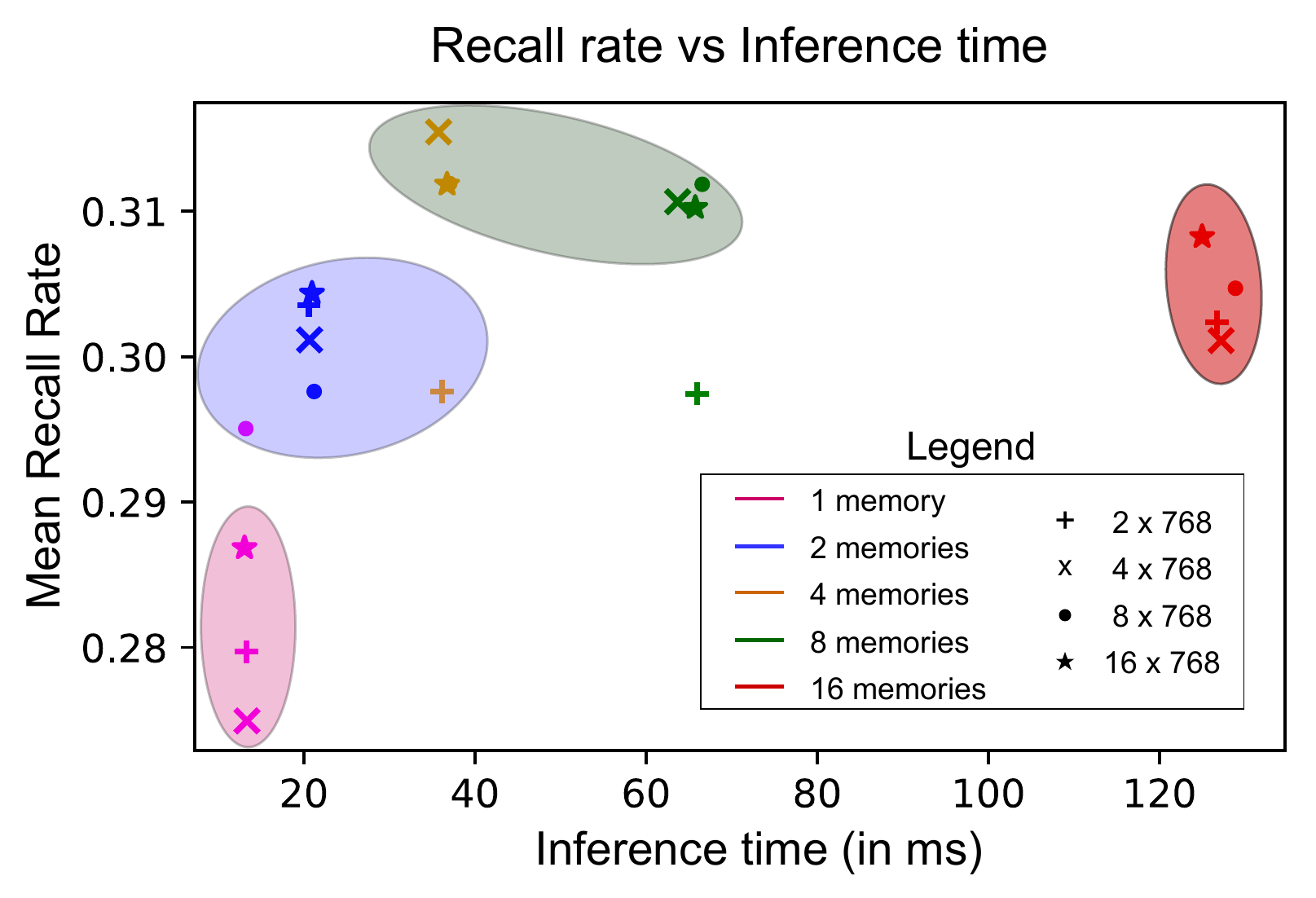}
  \caption{Scatter plot showing recall versus inference time for multiple memories in our CM-NTM for the Multi-turn Shoes dataset. The configurations belonging to the \textcolor{green}{green} cluster give the best recall overall, while having a high inference time. The configurations in the \textcolor{blue}{blue} cluster provide a suitable alternative with quicker inference at the cost of lower recall. The \textcolor{magenta}{magenta} and \textcolor{red}{red} clusters are undesirable configurations due to poor performance and long inference time, respectively.}
  \label{fig: scat_plot}
  \vspace{-15pt}
\end{figure}

\noindent\textbf{Inference time with multiple memories} - In this experiment, we study the performance of FashionNTM in terms of the mean of R@$5$ and R@$8$ recall rates along with the time taken to process one multi-turn transaction. For each $C$, we evaluate the performance across four different memory sizes. As seen from \Cref{fig: scat_plot}, the inference time increases with additional memories. Configurations in the green and blue clusters are desirable, as they provide a good trade-off between recall performance and computation time, while the pink and red clusters are undesirable due to poor performance and longer inference time, respectively.
\section{Conclusion and Future Work}

In this paper, we presented a novel cascaded Neural Turing Machine-based approach, called FashionNTM, for multi-turn feedback-based fashion image retrieval. Multiple memories in our model allow it to effectively retain and recall a number of complex relationships across transactions in the multi-turn setting, and multiple controllers help in assigning relative importance to each feature stored in the memory. This aids in attending to different parts of the input at different levels, thus leading to better performance across datasets. We also performed extensive experiments to compare our performance with baselines and previous state-of-the-art and observed that our multi-memory model significantly outperforms previous works \cite{yuan2021conversational}, with up to $50.5\%$ relative improvement on Multi-turn FashionIQ, and by $12.6\%$ on the multi-turn Shoes dataset, which we created in this work. We further demonstrated that our model can generalize beyond the trained setting to dynamically interact with real-world users to retrieve meaningful final product images. Finally, a user preference study reveals that our model is preferred by human participants on an average $83.1\%$ more as compared to other multi-turn methods. 

Despite promising results, there are a few limitations that make multi-turn image retrieval a hard problem to solve. Firstly, there is dearth of high quality and diverse multi-turn image retrieval datasets in the fashion domain which hinders comprehensive studies in this field.  Additionally, deploying current approaches to real-world scenarios (\eg, virtual private assistants) becomes a challenge due to computational requirements. Lastly, selecting the right configuration for different components such as memory size, number of memories, \etc. in a memory-based network is not a trivial task and depends a lot on the use case. Nevertheless, for future work, our approach could be further extended to non-fashion domains where multi-turn feedback-based information retrieval solutions are required.



{\small
\bibliographystyle{ieee_fullname}
\bibliography{egbib}

\begin{thebibliography}{10}\itemsep=-1pt

\bibitem{ewma}
{Moving Average}.
\newblock \url{https://en.wikipedia.org/wiki/Moving average}.

\bibitem{arandjelovic2016netvlad}
Relja Arandjelovic, Petr Gronat, Akihiko Torii, Tomas Pajdla, and Josef Sivic.
\newblock Netvlad: Cnn architecture for weakly supervised place recognition.
\newblock In {\em Proceedings of the IEEE conference on computer vision and
  pattern recognition}, pages 5297--5307, 2016.

\bibitem{9857242}
Alberto Baldrati, Marco Bertini, Tiberio Uricchio, and Alberto Del~Bimbo.
\newblock Conditioned and composed image retrieval combining and partially
  fine-tuning clip-based features.
\newblock In {\em 2022 IEEE/CVF Conference on Computer Vision and Pattern
  Recognition Workshops (CVPRW)}, pages 4955--4964, 2022.

\bibitem{Baldrati_2022_CVPR}
Alberto Baldrati, Marco Bertini, Tiberio Uricchio, and Alberto Del~Bimbo.
\newblock Effective conditioned and composed image retrieval combining
  clip-based features.
\newblock In {\em Proceedings of the IEEE/CVF Conference on Computer Vision and
  Pattern Recognition (CVPR)}, pages 21466--21474, June 2022.

\bibitem{berg2010automatic}
Tamara~L Berg, Alexander~C Berg, and Jonathan Shih.
\newblock Automatic attribute discovery and characterization from noisy web
  data.
\newblock In {\em European Conference on Computer Vision}, pages 663--676.
  Springer, 2010.

\bibitem{chen2020learning}
Yanbei Chen and Loris Bazzani.
\newblock Learning joint visual semantic matching embeddings for
  language-guided retrieval.
\newblock In {\em European Conference on Computer Vision}, pages 136--152.
  Springer, 2020.

\bibitem{Chen_2020_CVPR}
Yanbei Chen, Shaogang Gong, and Loris Bazzani.
\newblock Image search with text feedback by visiolinguistic attention
  learning.
\newblock In {\em Proceedings of the IEEE/CVF Conference on Computer Vision and
  Pattern Recognition (CVPR)}, June 2020.

\bibitem{cheng2022xmem}
Ho~Kei Cheng and Alexander~G Schwing.
\newblock Xmem: Long-term video object segmentation with an atkinson-shiffrin
  memory model.
\newblock In {\em European Conference on Computer Vision}, pages 640--658.
  Springer, 2022.

\bibitem{cho2014properties}
Kyunghyun Cho, Bart Van~Merri{\"e}nboer, Dzmitry Bahdanau, and Yoshua Bengio.
\newblock On the properties of neural machine translation: Encoder-decoder
  approaches.
\newblock {\em arXiv preprint arXiv:1409.1259}, 2014.

\bibitem{chopra2005learning}
Sumit Chopra, Raia Hadsell, and Yann LeCun.
\newblock Learning a similarity metric discriminatively, with application to
  face verification.
\newblock In {\em 2005 IEEE Computer Society Conference on Computer Vision and
  Pattern Recognition (CVPR'05)}, volume~1, pages 539--546. IEEE, 2005.

\bibitem{Datta_2022_CVPR}
Samyak Datta, Sameer Dharur, Vincent Cartillier, Ruta Desai, Mukul Khanna,
  Dhruv Batra, and Devi Parikh.
\newblock Episodic memory question answering.
\newblock In {\em Proceedings of the IEEE/CVF Conference on Computer Vision and
  Pattern Recognition (CVPR)}, pages 19119--19128, June 2022.

\bibitem{devlin2018bert}
Jacob Devlin, Ming-Wei Chang, Kenton Lee, and Kristina Toutanova.
\newblock Bert: Pre-training of deep bidirectional transformers for language
  understanding.
\newblock {\em arXiv preprint arXiv:1810.04805}, 2018.

\bibitem{fan2019heterogeneous}
Chenyou Fan, Xiaofan Zhang, Shu Zhang, Wensheng Wang, Chi Zhang, and Heng
  Huang.
\newblock Heterogeneous memory enhanced multimodal attention model for video
  question answering.
\newblock In {\em Proceedings of the IEEE/CVF conference on computer vision and
  pattern recognition}, pages 1999--2007, 2019.

\bibitem{Goenka_2022_CVPR}
Sonam Goenka, Zhaoheng Zheng, Ayush Jaiswal, Rakesh Chada, Yue Wu, Varsha
  Hedau, and Pradeep Natarajan.
\newblock Fashionvlp: Vision language transformer for fashion retrieval with
  feedback.
\newblock In {\em Proceedings of the IEEE/CVF Conference on Computer Vision and
  Pattern Recognition (CVPR)}, pages 14105--14115, June 2022.

\bibitem{gordo2016deep}
Albert Gordo, Jon Almaz{\'a}n, Jerome Revaud, and Diane Larlus.
\newblock Deep image retrieval: Learning global representations for image
  search.
\newblock In {\em European conference on computer vision}, pages 241--257.
  Springer, 2016.

\bibitem{graves2014neural}
Alex Graves, Greg Wayne, and Ivo Danihelka.
\newblock Neural turing machines.
\newblock {\em arXiv preprint arXiv:1410.5401}, 2014.

\bibitem{guo2018dialog}
Xiaoxiao Guo, Hui Wu, Yu Cheng, Steven Rennie, Gerald Tesauro, and Rogerio
  Feris.
\newblock Dialog-based interactive image retrieval.
\newblock {\em Advances in neural information processing systems}, 31, 2018.

\bibitem{han2017automatic}
Xintong Han, Zuxuan Wu, Phoenix~X Huang, Xiao Zhang, Menglong Zhu, Yuan Li,
  Yang Zhao, and Larry~S Davis.
\newblock Automatic spatially-aware fashion concept discovery.
\newblock In {\em Proceedings of the IEEE international conference on computer
  vision}, pages 1463--1471, 2017.

\bibitem{he2016deep}
Kaiming He, Xiangyu Zhang, Shaoqing Ren, and Jian Sun.
\newblock Deep residual learning for image recognition.
\newblock In {\em Proceedings of the IEEE conference on computer vision and
  pattern recognition}, pages 770--778, 2016.

\bibitem{hochreiter1997long}
Sepp Hochreiter and J{\"u}rgen Schmidhuber.
\newblock Long short-term memory.
\newblock {\em Neural computation}, 9(8):1735--1780, 1997.

\bibitem{hori2019end}
Chiori Hori, Huda Alamri, Jue Wang, Gordon Wichern, Takaaki Hori, Anoop
  Cherian, Tim~K Marks, Vincent Cartillier, Raphael~Gontijo Lopes, Abhishek
  Das, et~al.
\newblock End-to-end audio visual scene-aware dialog using multimodal
  attention-based video features.
\newblock In {\em ICASSP 2019-2019 IEEE International Conference on Acoustics,
  Speech and Signal Processing (ICASSP)}, pages 2352--2356. IEEE, 2019.

\bibitem{hosseinzadeh2020composed}
Mehrdad Hosseinzadeh and Yang Wang.
\newblock Composed query image retrieval using locally bounded features.
\newblock In {\em Proceedings of the IEEE/CVF Conference on Computer Vision and
  Pattern Recognition}, pages 3596--3605, 2020.

\bibitem{hou2021disentanglement}
Yuxin Hou, Eleonora Vig, Michael Donoser, and Loris Bazzani.
\newblock Learning attribute-driven disentangled representations for
  interactive fashion retrieval.
\newblock In {\em The International Conference on Computer Vision (ICCV)},
  October 2021.

\bibitem{howard2017mobilenets}
Andrew~G Howard, Menglong Zhu, Bo Chen, Dmitry Kalenichenko, Weijun Wang,
  Tobias Weyand, Marco Andreetto, and Hartwig Adam.
\newblock Mobilenets: Efficient convolutional neural networks for mobile vision
  applications.
\newblock {\em arXiv preprint arXiv:1704.04861}, 2017.

\bibitem{jegou2008hamming}
Herve Jegou, Matthijs Douze, and Cordelia Schmid.
\newblock Hamming embedding and weak geometric consistency for large scale
  image search.
\newblock In {\em European conference on computer vision}, pages 304--317.
  Springer, 2008.

\bibitem{ji2022lamemo}
Haozhe Ji, Rongsheng Zhang, Zhenyu Yang, Zhipeng Hu, and Minlie Huang.
\newblock Lamemo: Language modeling with look-ahead memory.
\newblock {\em arXiv preprint arXiv:2204.07341}, 2022.

\bibitem{jordan1997serial}
Michael~I Jordan.
\newblock Serial order: A parallel distributed processing approach.
\newblock In {\em Advances in psychology}, volume 121, pages 471--495.
  Elsevier, 1997.

\bibitem{clemkoantm2019}
Clement Joudet.
\newblock ntm - neural turing machines in pytorch.
\newblock \url{https://github.com/clemkoa/ntm}, 2019.

\bibitem{kingma2014adam}
Diederik~P Kingma and Jimmy Ba.
\newblock Adam: A method for stochastic optimization.
\newblock {\em arXiv preprint arXiv:1412.6980}, 2014.

\bibitem{le2020bist}
Hung Le, Doyen Sahoo, Nancy~F Chen, and Steven~CH Hoi.
\newblock Bist: Bi-directional spatio-temporal reasoning for video-grounded
  dialogues.
\newblock {\em arXiv preprint arXiv:2010.10095}, 2020.

\bibitem{Lee_2021_CVPR}
Seungmin Lee, Dongwan Kim, and Bohyung Han.
\newblock Cosmo: Content-style modulation for image retrieval with text
  feedback.
\newblock In {\em Proceedings of the IEEE/CVF Conference on Computer Vision and
  Pattern Recognition (CVPR)}, pages 802--812, June 2021.

\bibitem{li2011text}
Wen Li, Lixin Duan, Dong Xu, and Ivor Wai-Hung Tsang.
\newblock Text-based image retrieval using progressive multi-instance learning.
\newblock In {\em 2011 international conference on computer vision}, pages
  2049--2055. IEEE, 2011.

\bibitem{lin2019entropy}
Kuan-Yen Lin, Chao-Chun Hsu, Yun-Nung Chen, and Lun-Wei Ku.
\newblock Entropy-enhanced multimodal attention model for scene-aware dialogue
  generation.
\newblock {\em arXiv preprint arXiv:1908.08191}, 2019.

\bibitem{lin2017feature}
Tsung-Yi Lin, Piotr Doll{\'a}r, Ross Girshick, Kaiming He, Bharath Hariharan,
  and Serge Belongie.
\newblock Feature pyramid networks for object detection.
\newblock In {\em Proceedings of the IEEE conference on computer vision and
  pattern recognition}, pages 2117--2125, 2017.

\bibitem{liu2016deepfashion}
Ziwei Liu, Ping Luo, Shi Qiu, Xiaogang Wang, and Xiaoou Tang.
\newblock Deepfashion: Powering robust clothes recognition and retrieval with
  rich annotations.
\newblock In {\em Proceedings of the IEEE conference on computer vision and
  pattern recognition}, pages 1096--1104, 2016.

\bibitem{liu2021image}
Zheyuan Liu, Cristian Rodriguez-Opazo, Damien Teney, and Stephen Gould.
\newblock Image retrieval on real-life images with pre-trained
  vision-and-language models.
\newblock In {\em Proceedings of the IEEE/CVF International Conference on
  Computer Vision}, pages 2125--2134, 2021.

\bibitem{lu2020video}
Xiankai Lu, Wenguan Wang, Martin Danelljan, Tianfei Zhou, Jianbing Shen, and
  Luc~Van Gool.
\newblock Video object segmentation with episodic graph memory networks.
\newblock In {\em European Conference on Computer Vision}, pages 661--679.
  Springer, 2020.

\bibitem{https://doi.org/10.48550/arxiv.2109.00301}
Pedro~Henrique Martins, Zita Marinho, and André F.~T. Martins.
\newblock $\infty$-former: Infinite memory transformer, 2021.

\bibitem{masi2018deep}
Iacopo Masi, Yue Wu, Tal Hassner, and Prem Natarajan.
\newblock Deep face recognition: A survey.
\newblock In {\em 2018 31st SIBGRAPI conference on graphics, patterns and
  images (SIBGRAPI)}, pages 471--478. IEEE, 2018.

\bibitem{mirchandani-etal-2022-fad}
Suvir Mirchandani et~al.
\newblock {F}a{D}-{VLP}: Fashion vision-and-language pre-training towards
  unified retrieval and captioning.
\newblock In {\em Proceedings of the 2022 Conference on Empirical Methods in
  Natural Language Processing}, pages 10484--10497, Dec. 2022.

\bibitem{noh2017large}
Hyeonwoo Noh, Andre Araujo, Jack Sim, Tobias Weyand, and Bohyung Han.
\newblock Large-scale image retrieval with attentive deep local features.
\newblock In {\em Proceedings of the IEEE international conference on computer
  vision}, pages 3456--3465, 2017.

\bibitem{pham2022video}
Hoang-Anh Pham, Thao~Minh Le, Vuong Le, Tu~Minh Phuong, and Truyen Tran.
\newblock Video dialog as conversation about objects living in space-time.
\newblock {\em arXiv preprint arXiv:2207.03656}, 2022.

\bibitem{10.1007/978-3-319-46448-0_1}
Filip Radenovi{\'{c}}, Giorgos Tolias, and Ond{\v{r}}ej Chum.
\newblock Cnn image retrieval learns from bow: Unsupervised fine-tuning with
  hard examples.
\newblock In {\em Computer Vision -- ECCV 2016}, pages 3--20, 2016.

\bibitem{radenovic2018deep}
Filip Radenovic, Giorgos Tolias, and Ondrej Chum.
\newblock Deep shape matching.
\newblock In {\em Proceedings of the european conference on computer vision
  (eccv)}, pages 751--767, 2018.

\bibitem{radford2021learning}
Alec Radford, Jong~Wook Kim, Chris Hallacy, Aditya Ramesh, Gabriel Goh,
  Sandhini Agarwal, Girish Sastry, Amanda Askell, Pamela Mishkin, Jack Clark,
  et~al.
\newblock Learning transferable visual models from natural language
  supervision.
\newblock In {\em International Conference on Machine Learning}, pages
  8748--8763. PMLR, 2021.

\bibitem{rumelhart1985learning}
David~E Rumelhart, Geoffrey~E Hinton, and Ronald~J Williams.
\newblock Learning internal representations by error propagation.
\newblock Technical report, California Univ San Diego La Jolla Inst for
  Cognitive Science, 1985.

\bibitem{sandler2022fine}
Mark Sandler, Andrey Zhmoginov, Max Vladymyrov, and Andrew Jackson.
\newblock Fine-tuning image transformers using learnable memory.
\newblock In {\em Proceedings of the IEEE/CVF Conference on Computer Vision and
  Pattern Recognition}, pages 12155--12164, 2022.

\bibitem{schroff2015facenet}
Florian Schroff, Dmitry Kalenichenko, and James Philbin.
\newblock Facenet: A unified embedding for face recognition and clustering.
\newblock In {\em Proceedings of the IEEE conference on computer vision and
  pattern recognition}, pages 815--823, 2015.

\bibitem{650093}
M. Schuster and K.K. Paliwal.
\newblock Bidirectional recurrent neural networks.
\newblock {\em IEEE Transactions on Signal Processing}, 45(11):2673--2681,
  1997.

\bibitem{seong2020kernelized}
Hongje Seong, Junhyuk Hyun, and Euntai Kim.
\newblock Kernelized memory network for video object segmentation.
\newblock In {\em European Conference on Computer Vision}, pages 629--645.
  Springer, 2020.

\bibitem{sukhbaatar2015end}
Sainbayar Sukhbaatar, Jason Weston, Rob Fergus, et~al.
\newblock End-to-end memory networks.
\newblock {\em Advances in neural information processing systems}, 28, 2015.

\bibitem{turk1991eigenfaces}
Matthew Turk and Alex Pentland.
\newblock Eigenfaces for recognition.
\newblock {\em Journal of cognitive neuroscience}, 3(1):71--86, 1991.

\bibitem{NIPS2017_3f5ee243}
Ashish Vaswani, Noam Shazeer, Niki Parmar, Jakob Uszkoreit, Llion Jones,
  Aidan~N Gomez, \L~ukasz Kaiser, and Illia Polosukhin.
\newblock Attention is all you need.
\newblock In {\em Advances in Neural Information Processing Systems}, 2017.

\bibitem{vo2019composing}
Nam Vo, Lu Jiang, Chen Sun, Kevin Murphy, Li-Jia Li, Li Fei-Fei, and James
  Hays.
\newblock Composing text and image for image retrieval-an empirical odyssey.
\newblock In {\em CVPR}, 2019.

\bibitem{weston2014memory}
Jason Weston, Sumit Chopra, and Antoine Bordes.
\newblock Memory networks.
\newblock {\em arXiv preprint arXiv:1410.3916}, 2014.

\bibitem{Wu_2021_CVPR}
Hui Wu, Yupeng Gao, Xiaoxiao Guo, Ziad Al-Halah, Steven Rennie, Kristen
  Grauman, and Rogerio Feris.
\newblock Fashion iq: A new dataset towards retrieving images by natural
  language feedback.
\newblock In {\em Proceedings of the IEEE/CVF Conference on Computer Vision and
  Pattern Recognition (CVPR)}, pages 11307--11317, June 2021.

\bibitem{wu2020memformer}
Qingyang Wu, Zhenzhong Lan, Jing Gu, and Zhou Yu.
\newblock Memformer: The memory-augmented transformer.
\newblock {\em arXiv preprint arXiv:2010.06891}, 2020.

\bibitem{xie2021efficient}
Haozhe Xie, Hongxun Yao, Shangchen Zhou, Shengping Zhang, and Wenxiu Sun.
\newblock Efficient regional memory network for video object segmentation.
\newblock In {\em CVPR}, 2021.

\bibitem{XuewenECCV20Fashion}
Xuewen Yang, Heming Zhang, Di Jin, Yingru Liu, Chi-Hao Wu, Jianchao Tan,
  Dongliang Xie, Jue Wang, and Xin Wang.
\newblock Fashion captioning: Towards generating accurate descriptions with
  semantic rewards.
\newblock In {\em ECCV}, 2020.

\bibitem{yuan2021conversational}
Yifei Yuan and Wai Lam.
\newblock Conversational fashion image retrieval via multiturn natural language
  feedback.
\newblock In {\em Proceedings of the 44th International ACM SIGIR Conference on
  Research and Development in Information Retrieval}, pages 839--848, 2021.

\bibitem{zhang2005user}
Chen Zhang, Joyce~Y Chai, and Rong Jin.
\newblock User term feedback in interactive text-based image retrieval.
\newblock In {\em Proceedings of the 28th annual international ACM SIGIR
  conference on Research and development in information retrieval}, pages
  51--58, 2005.

\bibitem{zhang2022text}
Ruiyi Zhang, Tong Yu, Yilin Shen, and Hongxia Jin.
\newblock Text-based interactive recommendation via offline reinforcement
  learning.
\newblock In {\em Proceedings of the AAAI Conference on Artificial
  Intelligence}, volume~36, pages 11694--11702, 2022.

\bibitem{zhang2019text}
Ruiyi Zhang, Tong Yu, Yilin Shen, Hongxia Jin, and Changyou Chen.
\newblock Text-based interactive recommendation via constraint-augmented
  reinforcement learning.
\newblock {\em Advances in neural information processing systems}, 32, 2019.

\end{thebibliography}
}

\clearpage
\newpage
\section*{Supplementary Material}
\textbf{Overview} - In this supplementary material, we present some additional illustrations, and extended experiments from those we have in the main paper. Specifically, we show the following:
\begin{itemize}
    \item In the main paper, we showed the ablation study on Shoes dataset. Here, we show similar results for the FashionIQ dataset.
    \item We provide some illustrations for the multi-turn version of the existing Shoes \cite{berg2010automatic} dataset that we contributed to in our work. We further show how the annotations in this dataset are more specific, and consistent than multi-turn FashionIQ \cite{yuan2021conversational}.
    \item We also add some details, and illustrations of the user study experiment that we conducted.
\end{itemize}

\section{Ablation Study on FashionIQ}
In the main paper, we conduced ablation study on the multi-turn Shoes dataset. Here, we show results of similar experiments on the multi-turn FashionIQ dataset. In \Cref{tab:num_mem_ablation_supp}, we show an ablation over different number of memories of our proposed FashionNTM approach, by fixing the memory size to $8 \times 768$. 

\begin{table}[h]
\scriptsize
\centering
\caption{Different number of memories for the proposed approach by fixing the memory size to $8 \times 768$ for FashionIQ dataset. We select the mean value for comparison and pick the best one.}
\label{tab:num_mem_ablation_supp}
\begin{tabular}{c|c|ccc|c}\toprule
\multirow{2}{*}{\textbf{Model}} & \textbf{Number of}& \multirow{2}{*}{\textbf{R}@$\mathbf{5}$} & \multirow{2}{*}{\textbf{R}@$\mathbf{8}$} & \multirow{2}{*}{\textbf{Mean}} & $\mathbf{\%}$\\
&\textbf{memories (C)}&&&&\textbf{increase}\\
\hline
ST+v-NTM & $1$ & $44.8$ & $50.0$ & $47.4$ & -\\
\hline
\multirow{4}{*}{FashionNTM} &$2$ & $45.0$ & $50.1$ & $47.5$ & $0.2$\\
&$4$ & $45.3$ & $50.3$ & $47.8$ & $0.8$\\
&$8$ & $\textbf{45.7}$ & $\textbf{50.4}$ & $\textbf{48.1}$ & $\textbf{1.5}$\\
&$16$ & $44.9$ & $50.0$ & $47.4$ & $0$\\
\bottomrule
\end{tabular}
\end{table}

This is analogous to \Cref{tab:num_mem_ablation} in the main paper, where we conducted a similar experiment for multi-turn Shoes dataset. We observe a similar trend here too, as the performance gradually increases with more cascaded stages, before peaking at $C=8$, and then reduces as the memory network becomes larger at $C=16$. An interesting observation in \Cref{tab:num_mem_ablation_supp} is that the relative improvement is not as high as that for Shoes dataset (see \Cref{tab:num_mem_ablation} of main paper). This is primarily due to the differences between the annotation quality of the two datasets.

\begin{figure}[ht]
    \centering
  \includegraphics[width=0.9\linewidth,trim={0.5cm 0.5cm 0 0.5cm},clip]{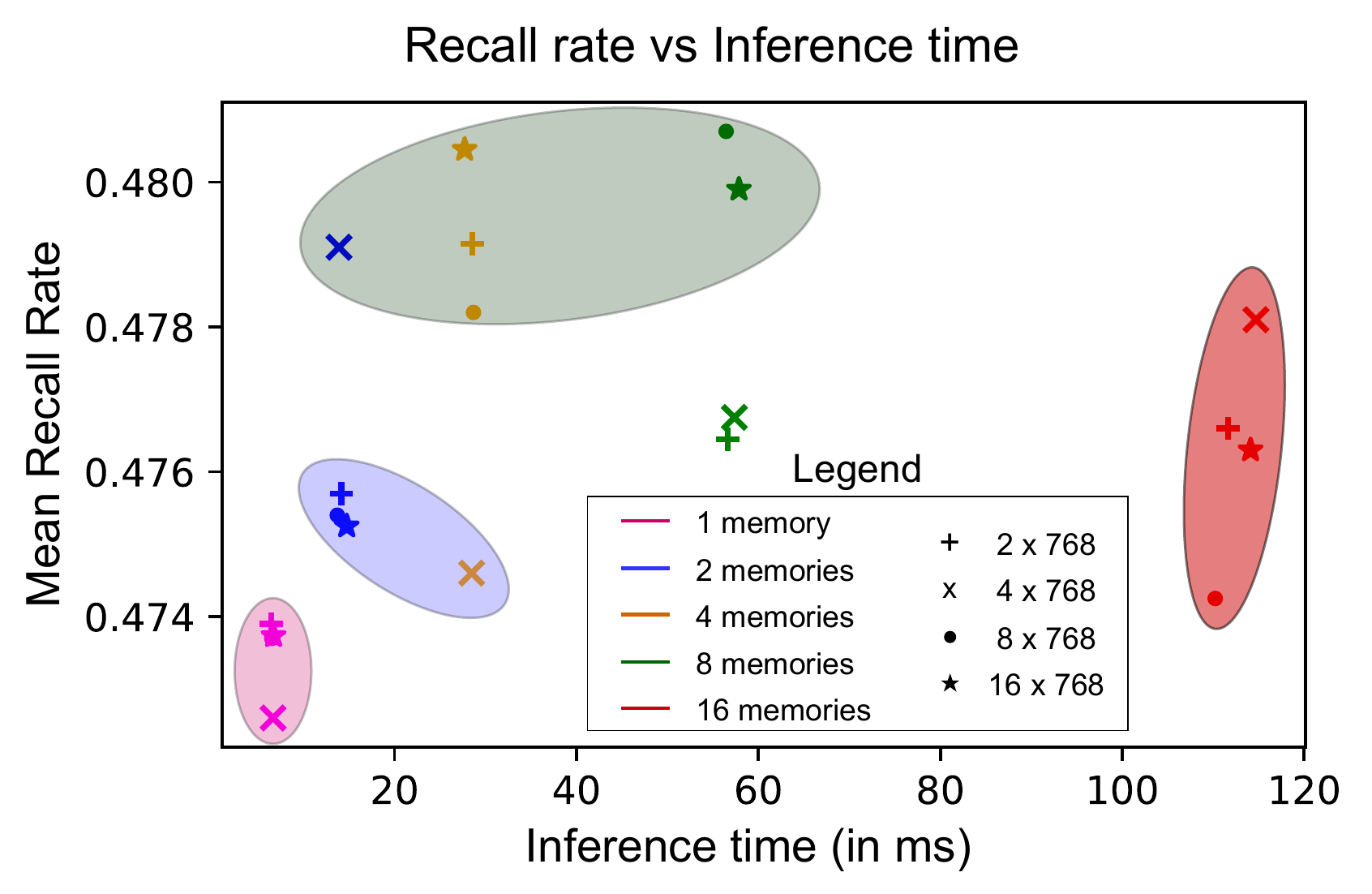}
  \caption{Scatter plot showing recall versus inference time for multiple memories in our CM-NTM for the Multi-turn FashionIQ dataset. The configurations belonging to the \textcolor{green}{green} cluster give the best recall overall, while having a high inference time. The configurations in the \textcolor{blue}{blue} cluster provide a suitable alternative with quicker inference at the cost of lower recall. The \textcolor{magenta}{magenta} and \textcolor{red}{red} clusters are undesirable configurations due to poor performance and long inference time, respectively.}
  \label{fig: scat_plot_supp}
\end{figure}

\begin{figure*}
\centering
\begin{minipage}[c]{0.49\linewidth}
  \includegraphics[width=\linewidth]{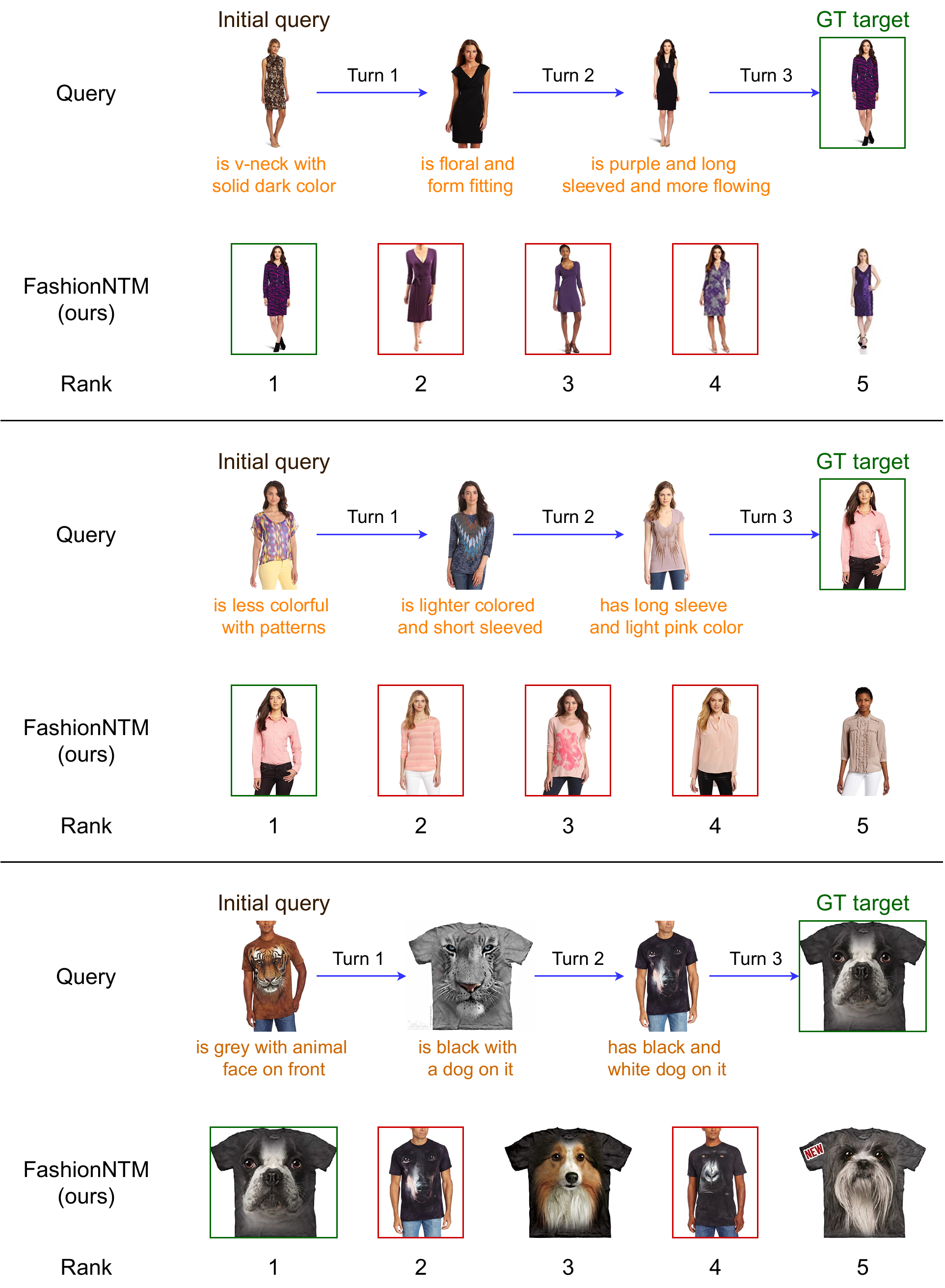}
  \caption{Top-5 retrievals of our method on FashionIQ dataset. The images with \textcolor{green}{green} bounding box are ground-truth target. The images with \textcolor{red}{red} bounding box have similar properties as those of the GT, but are not marked. This shows that annotations in FashionIQ dataset are too generic, thus corresponidng to multiple images.}
  \label{fig: fiq_topk_supp}
\end{minipage}
\hfill
\begin{minipage}[c]{0.49\linewidth}
  \includegraphics[width=\linewidth]{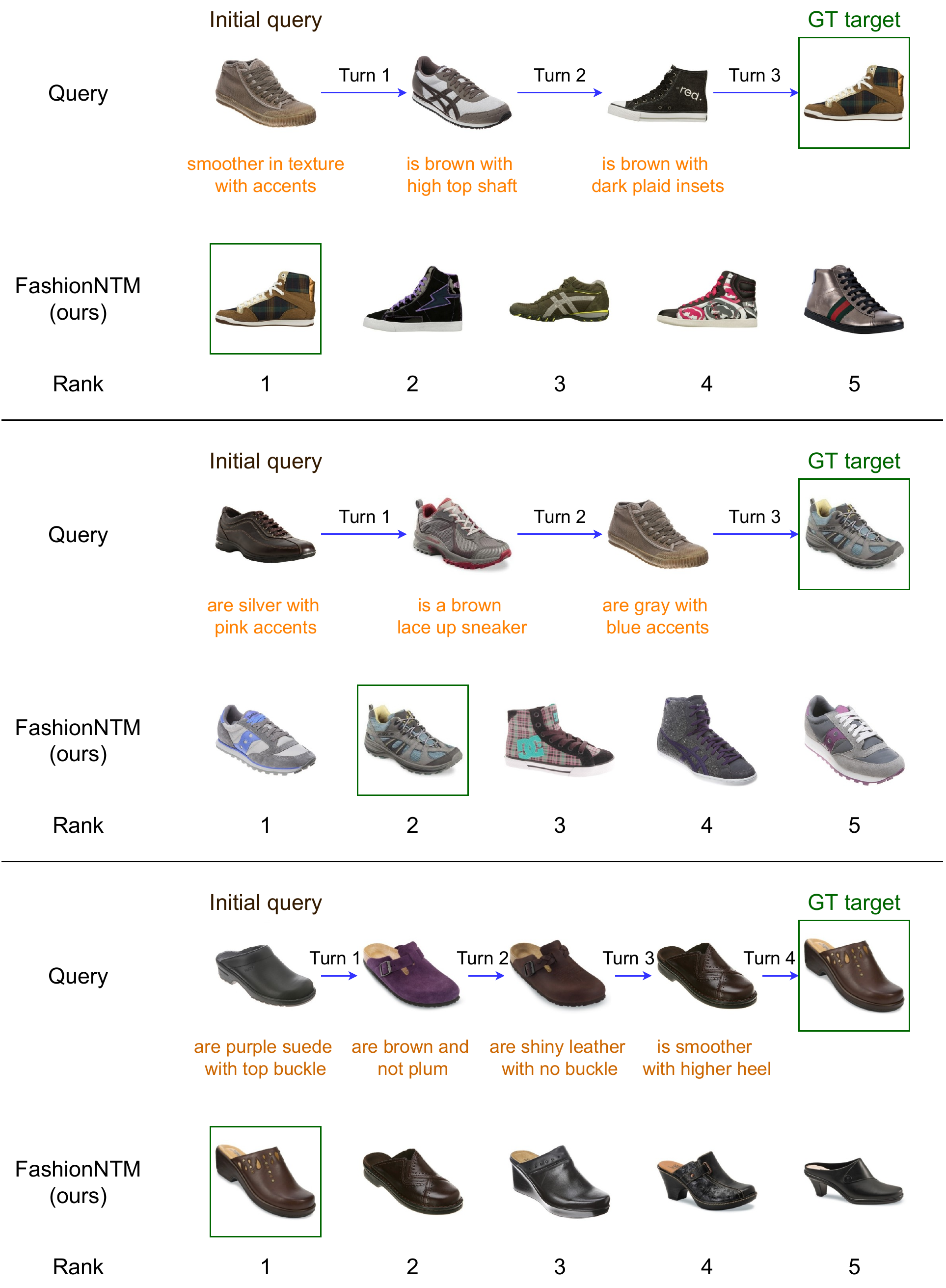}
  \caption{Top-5 retrievals of our method on Shoes dataset. The images with \textcolor{green}{green} bounding box correspond to ground-truth target. Due to more consistent and accurate descriptions, the feedback text in this dataset typically corresponds to only one image, thereby reducing ambiguity in retrieval.}
  \label{fig: shoes_topk_supp}
\end{minipage}
\end{figure*}

In \Cref{fig: scat_plot_supp}, we plot the relative performance of all the experiments that we conducted on the multi-turn FashionIQ dataset. This is analogous to \Cref{fig: scat_plot} of the main paper. We observe the same type of trend for both datasets -- inference time increases as we add more memories. Configurations in the green and blue clusters are desirable, as they provide a good trade-off between recall performance and computation time, while the pink and red clusters are undesirable due to poor performance and longer inference time, respectively.

\begin{figure*}[t]
    \centering
  \includegraphics[width=\linewidth]{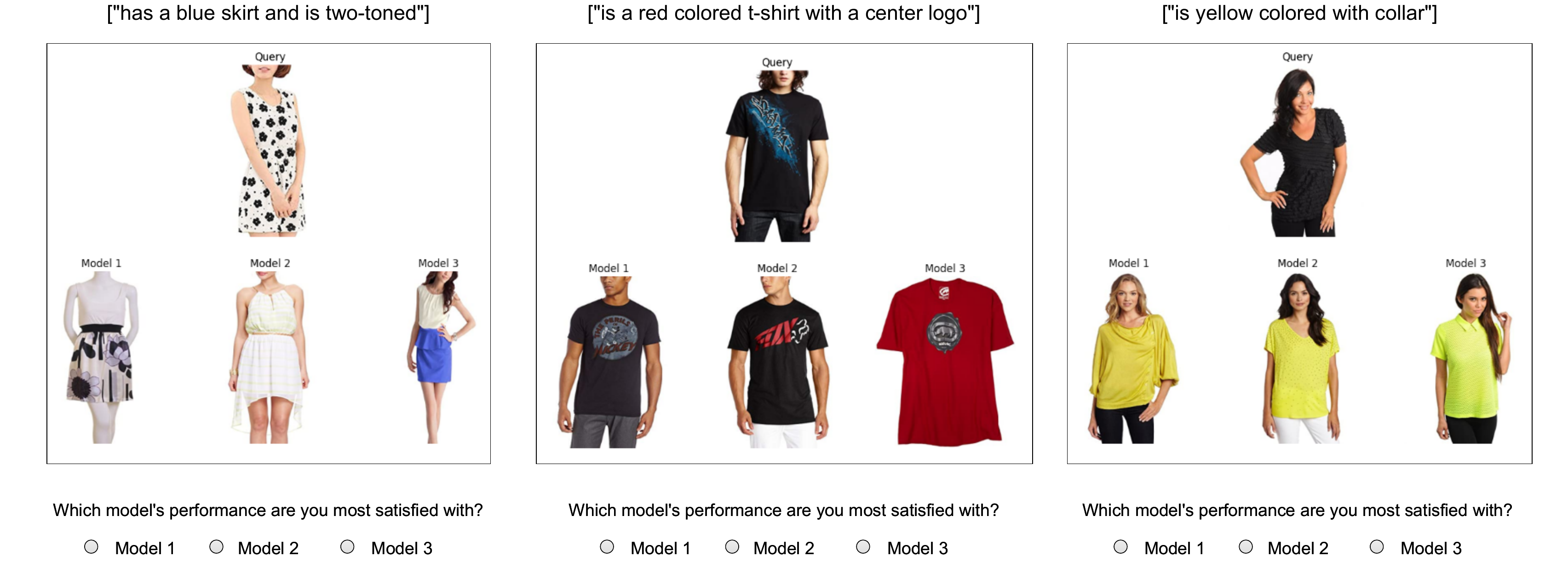}
  \caption{User interface shown to the participants for the human preference study. Here, we show three examples of top-1 retrievals of 3 best multi-turn models on the multi-turn FashionIQ dataset. Model $1$ corresponds to $\mathtt{ST+EWMA}$, while Model $2$ to $\mathtt{ST+LSTM}$. Finally, our proposed $\mathtt{FashionNTM}$ model is Model $3$. For fairness, we did not disclose the identity of the models to the participants.}
  \label{fig: user_study_supp}
  \vspace{-10pt}
\end{figure*}

\section{Difference between FashionIQ and Shoes} \label{sec: diff_data}
During our experiments, we observed that descriptions in the multi-turn FashionIQ \cite{yuan2021conversational} dataset feedbacks are very generic, and therefore can correspond to multiple target images. An example of this is shown in \Cref{fig: fiq_topk_supp}, where we demonstrate additional examples of the top-5 retrieval experiments, that we conducted in \Cref{fig: topk_qual_fiq} of the main paper. As seen from all three cases, there are \textit{multiple} target images, that match the description as provided by the feedback (e.g. in the top row, all $5$ retrieved images are ``purple and flowing", and $4$ of them are also ``long sleeved"). However, only one of them is labeled as the correct target (ground-truth) for a particular transaction. This makes it difficult to evaluate multi-turn systems on this dataset, as the performance (recall rate) might be low, even though the requirements are satisfied.

In contrast, for the multi-turn Shoes dataset, feedback texts are more concise, and hence there are usually very few target images ($\approx1 \text{ or } 2$) that match the description given in the feedback. This makes the Shoes dataset more suitable for evaluating multi-turn systems. Top-5 retrieval results for multi-turn Shoes dataset, in addition to those in \Cref{fig: topk_qual_shoes} of the main paper, are shown in \Cref{fig: shoes_topk_supp}.

\section{User study details}

In the main paper, we showed the results of a user preference study conducted among $5$ human participants. In \Cref{fig: user_study_supp}, we show some examples of the actual interface that was shown to the users. Each participant was shown $45$ such images, and asked a simple question about ``Which model's performance are you most satisfied with?" Based on their understanding, the users had to choose between the top images retrieval of $3$ models. To maintain fairness, we did not disclose the identity of the models to the participants. At the end of the study, we aggregated the votes given to the models by each user, and plotted it as a histogram as shown in \Cref{fig: rebuttal_plot} of the main paper.




\end{document}